\documentclass[acmlarge]{acmart}

\makeatletter
\newcommand{\myconfshort}{\acmConference@shortname}
\newcommand{\myconffull}{\acmConference@name}
\newcommand{\myconfdate}{\acmConference@date}
\newcommand{\myconfloc}{\acmConference@venue}
\AtBeginDocument{
  \fancypagestyle{firstpagestyle}{
    \fancyhead{}%
    \fancyfoot[C]{}%
  }
  \fancyhf{}
  \fancyhead[LO]{\@headfootfont\shorttitle}%
  \fancyhead[RE]{\@headfootfont\@shortauthors}%
  \fancyhead[LE]{\@headfootfont\footnotesize \myconfshort, \myconfdate, \myconfloc}%
  \fancyhead[RO]{\@headfootfont\footnotesize \myconfshort, \myconfdate, \myconfloc}%
  \fancyfoot[C]{}%
}
\makeatother
\acmBooktitle{\conffull\@ (\confshort), \confdate, \confloc}

\AtBeginDocument{%
  }

\acmYear{2026}
\acmDOI{10.1145/3805689.3806475}
\acmConference[FAccT '26]{The 2026 ACM Conference on Fairness, Accountability, and Transparency}{June 25--28, 2026}{Montreal, QC, Canada}

\acmBooktitle{The 2026 ACM Conference on Fairness, Accountability, and Transparency (FAccT '26), June 25--28, 2026, Montreal, QC, Canada}
\acmISBN{979-8-4007-2596-8/2026/06}

\usepackage{algorithm}
\usepackage[noend]{algpseudocode}
\usepackage{subcaption}
\usepackage{multirow}
\usepackage{xspace}
\usepackage{environ}
\usepackage{mathtools}
\usepackage[svgnames, x11names, table, dvipsnames]{xcolor}

\theoremstyle{plain}
\newtheorem{theorem}{Theorem}[section]
\newtheorem{lemma}[theorem]{Lemma}
\theoremstyle{definition}
\newtheorem{definition}[theorem]{Definition}

\newcommand{\paren}[1]{\left( #1 \right)}

\newcommand{\curly}[1]{\left\{ #1 \right\}}

\newcommand{\norm}[1]{\left\lVert #1 \right\rVert}

\DeclareMathOperator*{\E}{\mathbb{E}}
\DeclareMathOperator{\clip}{clip}

\newcommand{\cA}{\mathcal{A}}
\newcommand{\cB}{\mathcal{B}}

\newcommand{\cH}{\mathcal{H}}

\newcommand{\cN}{\mathcal{N}}
\newcommand{\cP}{\mathcal{P}}

\newcommand{\cX}{\mathcal{X}}
\newcommand{\cY}{\mathcal{Y}}
\newcommand{\cZ}{\mathcal{Z}}

\newcommand{\cPt}{\cP_t}

\newcommand{\g}[1]{g_{#1}}

\newcommand{\gbar}[1]{\bar{g}_{#1}}
\newcommand{\gbarclip}[1]{\bar{g}^{\text{clip}}_{#1}}
\newcommand{\gbardp}[1]{\bar{g}^{\text{DP}}_{#1}}

\newcommand{\edp}{(\varepsilon, \, \delta)}

\newcommand{\dpnoise}{\cN(0,\, \sigma^2 C^2 I_d)}
\newcommand{\dpnoisetwo}[1]{\cN\paren{0,\, \frac{\sigma^2 C^2}{|\cB_{#1}|^2} I_d}}
\newcommand{\dpnoisetwosl}[1]{\cN\paren{0,\, \paren{\sigma^2 C^2 / |\cB_{#1}|^2} I_d}}

\newcommand{\bac}{B_{\alpha, C}}
\newcommand{\sfactor}{\Gamma_{T}}
\newcommand{\xic}{\xi^c}
\newcommand{\xicmin}{\xi^c_{\mathrm{min}}}
\newcommand{\ximin}{\xi_{\mathrm{min}}}

\definecolor{burntorange}{rgb}{0.8, 0.33, 0.0}
\newcommand{\blue}[1]{\textcolor{RoyalPurple}{#1}}
\newcommand{\red}[1]{\textcolor{burntorange}{#1}}

\newcommand{\cifarten}{CIFAR-10\xspace}
\newcommand{\cifarh}{CIFAR-100\xspace}
\newcommand{\mnist}{MNIST\xspace}
\newcommand{\bm}{Bank Marketing\xspace}
\newcommand{\resume}{Resume\xspace}

\newcommand{\sgd}{SGD\xspace}
\newcommand{\dpsgd}{DP-SGD\xspace}
\newcommand{\dpx}{DP\xspace}
\newcommand{\aca}{ACA\xspace}

\begin{document}

\title{Crowding Out the Noise: Algorithmic Collective Action Under Differential Privacy}

\author{Rushabh Solanki}
\email{r7solank@uwaterloo.ca}
\affiliation{%
  \institution{University of Waterloo, Vector Institute}
  \city{Waterloo}
  \state{Ontario}
  \country{Canada}
}

\author{Meghana Bhange}
\email{meghana-shashikant.bhange.1@etsmtl.net}
\affiliation{%
  \institution{\'ETS Montr\'eal, Mila}
  \city{Montr\'eal}
  \state{Quebec}
  \country{Canada}
}

\author{Ulrich A\"ivodji}
\email{ulrich.aivodji@etsmtl.ca}
\affiliation{%
  \institution{\'ETS Montr\'eal, Mila}
  \city{Montr\'eal}
  \state{Quebec}
  \country{Canada}
}

\author{Elliot Creager}
\email{creager@uwaterloo.ca}
\affiliation{%
  \institution{University of Waterloo, Vector Institute}
  \city{Waterloo}
  \state{Ontario}
  \country{Canada}
}


\begin{abstract}
    The integration of AI into daily life has generated considerable attention and excitement, while also raising concerns about automating algorithmic harms and re-entrenching existing social inequities.
    While the responsible deployment of trustworthy AI systems is a worthy goal, there are many possible ways to realize it, from policy and regulation to improved algorithm design and evaluation.
    In fact, since AI trains on social data, there is even a possibility for everyday users, citizens, or workers to directly steer the AI system's behavior through \emph{Algorithmic Collective Action}, by deliberately modifying the data they share with a platform to drive its learning process in their favor.
    This paper considers how these grassroots efforts to influence AI interact with methods already used by AI firms and governments to improve model trustworthiness.
    In particular, we focus on the setting where the AI firm deploys a differentially private model, motivated by the growing regulatory focus on privacy and data protection.
    We investigate how the use of Differentially Private Stochastic Gradient Descent (\dpsgd) affects the collective's ability to influence the learning process.
    Our findings show that while differential privacy contributes to the protection of individual data, it introduces challenges for effective algorithmic collective action.
    We establish this trade-off formally by characterizing lower bounds on the success of algorithmic collective action under differential privacy as a function of the collective's size and the firm's privacy parameters.
    We then verify these trends experimentally by simulating collective action during the training of deep neural network classifiers across several datasets.
    Finally, we perform  a stylized economic analysis of privacy costs in order to integrate additional incentives at play for both parties, analyzing how factors like average utility and participation costs influence the formation of collectives under private training regimes.
\end{abstract}


\received{13 January 2026}

\maketitle

\section{Introduction}
\label{sec:introduction}

The rapid proliferation of AI systems across multiple domains has been propelled by the ability of AI firms to collect vast amounts of data for training purposes, sourced from public websites, users of the firm's products, and crowd workers.
By leveraging these large-scale datasets, the firms can train increasingly sophisticated models that not only improve their predictive capabilities but also expand the range of problems they can address.
Despite its advantages, the extensive use of personal data in training machine learning models has introduced pressing concerns about algorithmic harms, such as threats to privacy, exposure of sensitive information, and biased decision-making that can perpetuate social disparities.

In response to these concerns, various solutions have been proposed and implemented at different stages of the model development pipeline.
At the firm level, efforts towards building ``trustworthy AI" often involve fairness assessments, bias mitigation techniques, privacy auditing, and adversarial evaluations such as red teaming, with these efforts spanning multiple stages from data collection to model training and post-processing \cite{barocas2023fairnessml}.
However, implementing these techniques may introduce trade-offs with the firm's broader objectives of maximizing predictive performance and enhancing user engagement to generate more data.
On the other hand, several regional regulations, such as the General Data Protection Regulation (GDPR) \cite{eu2016gdpr}, the Personal Information Protection and Electronic Documents Act (PIPEDA) \cite{canada2000pipeda}, and the California Privacy Rights Act (CPRA) \cite{california2020cpra} establish baseline privacy protections; yet compliance with these laws alone does not guarantee socially responsible outcomes \cite{selbst2019fairness, utz2019uninformed}.
In parallel with organizational and regulatory measures, grassroots efforts of \emph{Algorithmic Collective Action} are taking shape \cite{hardt2024aca}, where users actively organize and contribute their data in a coordinated manner to strategically influence model behavior ``from below'' \cite{devrio2024building}.

Algorithmic collective action (\aca) \cite{hardt2024aca, olson1965collectiveaction} provides a principled framework for understanding how a group of individuals, through coordinated changes in their data, can impact the behavior of deployed models.
For instance, consider a scenario where job applicants subject to an automated skill-profiling algorithm coordinate to insert a shared keyword, such as a standardized formatting tag, in their resume.
By planting this signal, the collective aims to shift the model's learned associations, increasing the likelihood that the algorithm categorizes their profiles under a sought-after skill designation.
Prior work has provided theoretical insights under assumptions such as Bayes optimality, empirical risk minimization \cite{hardt2024aca}, or robust optimization \cite{bendov2024rolelearningaca}, offering an informed view of how these assumptions can affect the effectiveness of collective action on model behavior.
However, the interaction between the actions of coordinated users and privacy-preserving techniques employed by model owners remains largely unexplored.

In this paper, we investigate this intersection, focusing on Differential Privacy (\dpx), a widely used mathematical framework for protecting individual-level data through the injection of calibrated noise into the learning process.
In particular, we study the application of differential privacy in deep learning settings through Differentially Private Stochastic Gradient Descent (\dpsgd), a common approach for preserving privacy during model training.
Motivated by the strengthening of regulatory frameworks and the growing consumer demand for privacy guarantees, we seek to understand how differential privacy affects the learning algorithm's responsiveness to collective action and the likelihood of success of such interventions.

\begin{figure}[t]
    \centering
    \begin{subfigure}[b]{\linewidth}
        \begin{subfigure}[t]{0.24\linewidth}
            \includegraphics[width=\linewidth]{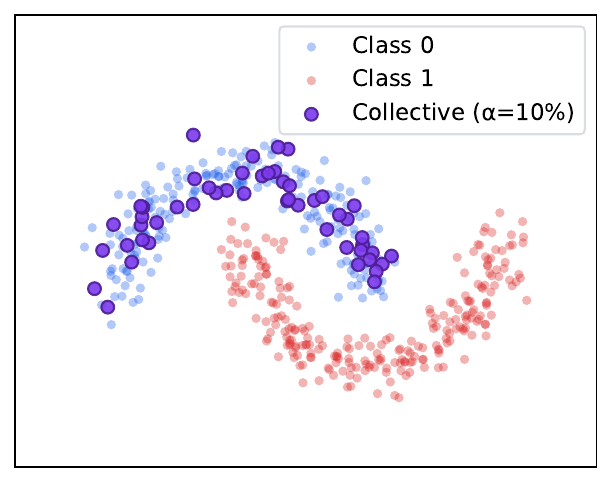}
            \caption{Collective's size $\alpha = 10\%$}
            \label{subfig:fig1a}
        \end{subfigure}
        \hfill
        \begin{subfigure}[t]{0.24\linewidth}
            \includegraphics[width=\linewidth]{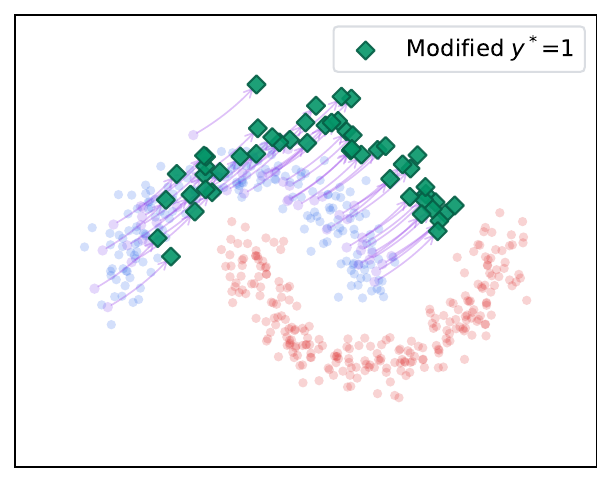}
            \caption{After collective action}
            \label{subfig:fig1b}
        \end{subfigure}
        \hfill
        \begin{subfigure}[t]{0.24\linewidth}
            \includegraphics[width=\linewidth]{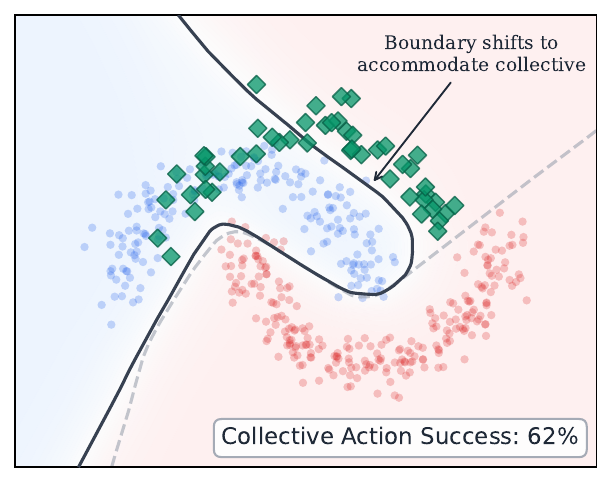}
            \caption{Trained under \sgd}
            \label{subfig:fig1c}
        \end{subfigure}
        \hfill
        \begin{subfigure}[t]{0.24\linewidth}
            \includegraphics[width=\linewidth]{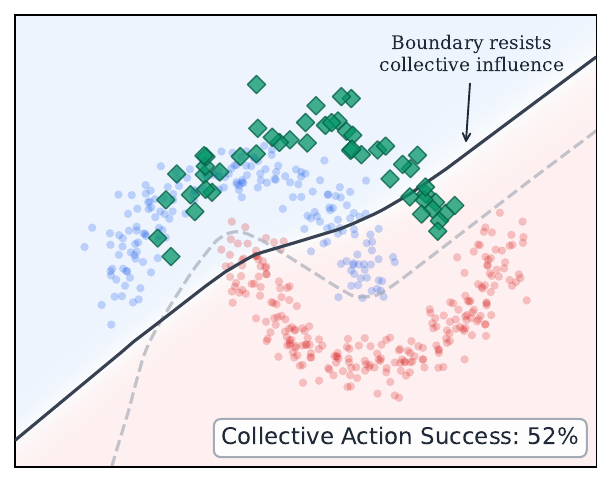}
            \caption{Trained under \dpsgd}
            \label{subfig:fig1d}
        \end{subfigure}
    \end{subfigure}
    \caption[
        Illustrative example of Algorithmic Collective Action under \sgd and \dpsgd.
    ]{
        Illustrative example of Algorithmic Collective Action under \sgd and \dpsgd. 
        \textbf{(a)} A group of size $\alpha=10\%$ participates in a collective action.
        \textbf{(b)} The collective aims to plant a signal to associate it with the desired label. To achieve this, they flip their labels to 1 and shift their features (the \emph{feature-label} strategy~\citep{hardt2024aca}).
        \textbf{(c)} Under standard \sgd, the decision boundary shifts to accommodate the collective's planted signal.
        \textbf{(d)} However, under \dpsgd, the boundary resists the collective action due to the addition of gradient noise and clipped gradients.
    }
    \Description{
        Illustrative example of Algorithmic Collective Action under \sgd and \dpsgd.
    }
    \label{fig:fig1}
\end{figure}

We propose a theory that examines the impact of differential privacy constraints on the effectiveness of the collective's actions on the firm's learning algorithm.
We operationalize this framework in practical deep learning scenarios and perform extensive experiments on multiple benchmark datasets, showing that, while differential privacy provides strong guarantees to protect individual data, it inadvertently reduces the collective's ability to coordinate and alter the behavior of the firm's model, as illustrated in Figure~\ref{fig:fig1}.
This work offers a new lens for the societal implications of using privacy-preserving techniques in machine learning, combining theoretical insight and empirical validation.

Our contributions are summarized as follows:
\begin{itemize}
    \item
    We identify and characterize a trade-off between Differential Privacy and Algorithmic Collective Action.
    Specifically, our theoretical model establishes lower bounds on the collective's success under differential privacy constraints, expressed in terms of the collective's size and the privacy parameters.
    \item
    We validate these theoretical findings through extensive experiments on multiple datasets, showing that differential privacy reduces the collective's ability to influence model behavior.
    \item 
    We perform a cost-benefit analysis to explore additional incentives for both the firm and the collective, highlighting how the collective can still achieve success even under stricter differential privacy constraints.
\end{itemize}

\section{Background}
\label{sec:background}

This section formally introduces algorithmic collective action and privacy-preserving training and defines the notation used throughout the paper.

\subsection{Collective Action}
\label{subsec:background-aca}

This work builds on the framework introduced by \citet{hardt2024aca}, which models the interaction between a firm's learning algorithm and a strategic collective of individuals.
The population is drawn from a base distribution $\cP_0$ over data points $z = (x, \, y) \in \cZ = \cX \times \cY$, where $\cX$ and $\cY$ denote the feature and label spaces.
Within the framework, a fraction $\alpha > 0$ of the population is controlled by the collective, which applies a possibly randomized modification strategy $h: \cZ \to \cZ$ chosen from a feasible set $\cH$.
Applying $h$ to data drawn from $\cP_0$ induces the collective's distribution $\cP^*$, and the firm's learning algorithm $\cA$ then encounters the following mixture data distribution $\cP$ defined as:
\begin{equation}
    \cP = \alpha \cP^* + (1 - \alpha) \cP_0.
    \label{eq:mixture-distribution}
\end{equation}
The firm's learning algorithm $\cA$ produces a classifier $f = \cA (\cP): \cX \to \cY$, trained on data drawn from $\cP$.

\paragraph{Planting a Signal}
We focus on a particular class of strategies in which a collective intentionally \emph{plants a signal} by modifying both features and labels of the data it controls, a setup referred to as the \emph{feature-label} strategy by \citet{hardt2024aca}.
The data are modified such that the classifier $f$ learns to associate the transformed version of the features with the chosen target label $y^{*}$.
The transformation is defined by a function $g: \cX \rightarrow \cX$, which results in the following strategy:
\begin{equation}
    h(x, \, y) = (g(x), \, y^{*}).
    \label{eq:feature-label-strategy}
\end{equation}
Contextualizing this within the automated skill-profiling scenario, $X$ denotes resume content and $Y$ the target skill categories.
The transformation $g$ corresponds to the collective's coordinated insertion of a formatting tag, coupling the modified input with a specific target label $y^*$, such as ``Python Developer''.

\paragraph{Critical Mass}
We are interested in determining the smallest collective size capable of achieving a desired level of success, which is referred to as the \emph{critical mass}.
Formally, for a given target success level $S^{*}$, the critical mass is the smallest value $\alpha$ such that the achieved success $S(\alpha) \geq S^{*}$.

\paragraph{Definition of Success}
While several learning-theoretic settings are explored in \citet{hardt2024aca}, this work focuses on characterizing the success criteria of the collective in the context of gradient-based optimization, where the learner selects a model from a parameterized family $\{ f_\theta \}_{\theta \in \Theta}$.
Let $\g{\cPt} (\theta_t) = \E_{z \sim \cPt} [\nabla \ell (\theta_t; z)]$ denote the expected gradient of the loss over the distribution $\cPt$, evaluated at the parameter $\theta_t \in \Theta$.
The learner then performs the following gradient descent update $\theta_{t + 1} = \theta_t - \eta \, \g{\cPt} (\theta_t)$.
By defining the target model $\theta^*$ that the collective desires to achieve by influencing the firm's model $\theta$, we measure the success of the collective after $t$ steps as $S_t(\alpha) = - \norm{\theta_{t} - \theta^*}$.

\paragraph{Collective's Strategy}
One natural way for the collective to steer the firm's model toward a target parameter $\theta^*$ is by influencing the overall gradient used by gradient descent so that it points, on average, in the desired direction.
Suppose at each update the training data comes from a mixture distribution defined in Equation~\eqref{eq:mixture-distribution}, the gradient of the loss under the mixture is $g_{\cP}(\theta_t) = (1 - \alpha) \, g_{\cP_0}(\theta_t) + \alpha \, g_{\cP^*}(\theta_t)$.
If the collective can choose $g_{\cP^*}(\theta)$ so that the mixture gradient equals a vector pointing from $\theta_t$ toward $\theta^*$, then gradient descent will move the parameter toward $\theta^*$. A convenient target for the mixture gradient is $g_{\cP}(\theta) = \xi(\theta_t) \cdot (\theta_t - \theta^*)$ for some scalar $\xi(\theta_t) > 0$.
The precise form of the required gradient is defined below.

\begin{definition}
    [Gradient-redirecting distribution from \citet{hardt2024aca}] \label{def:grad_redirect}
    Given an observed model $\theta$ and a target model $\theta^*$, the collective finds a \emph{gradient-redirecting distribution} $\cP'$ for $\theta$ where:
    \begin{equation}
        \g{\cP'}(\theta) = -\frac{1 - \alpha}{\alpha} \g{\cP_0}(\theta) + \xi (\theta - \theta^*),
        \label{eq:original-xi}
    \end{equation}
    for some $\xi \in \left(0, \tfrac{1}{\alpha \eta}\right)$. Once such a distribution is identified, we can sample modified data $z' \sim \cP'$ to guide the optimization process by setting $h(z) = z^\prime$.
\end{definition}

So long as the collective's data adheres to the gradient-redirecting distribution $\cP'$, the model's gradient updates on the mixture distribution will push it toward the collective's desired parameter $\theta^*$.
Intuitively, the gradient under the distribution $\cP'$ is composed of two functional terms.
Returning to our skill-profiling example, the first term reverses and rescales the original gradients $\g{\cP_0}(\theta)$, neutralizing what the model would have learned from standard resumes.
The second term is constructed so that, after gradient descent subtracts it, the net parameter update moves $\theta_t$ toward $\theta^*$, the model state that categorizes resumes containing the collective's shared keyword into their target skill.
The following theorem formalizes a lower bound on the success of the collective when applying the gradient-redirecting strategy.

\begin{theorem} 
    [Theorem 10 from \citet{hardt2024aca}]
    \label{theorem:hardt-theorem-10}
    Assume the collective can implement the gradient-redirecting strategy at all $\lambda \theta_0 + (1 - \lambda) \theta^*, \lambda \in [0, 1]$. Then, there exists $C(\alpha) > 0$ such that the success of the gradient-redirecting strategy after $T$ steps is lower bounded by,
    \begin{align*}
        S_{T}(\alpha) &\geq - \left( 1 - \eta C(\alpha) \right)^T
        \norm{\theta_0 - \theta^*}.
    \end{align*}
\end{theorem}
\noindent
Here, $ C(\alpha)$ is directly proportional to the collective's size $\alpha$.
As $\alpha$ increases (and consequently $C(\alpha)$), the lower bound on the collective's success also increases.
This result further implies that the collective can attain any desired model $\theta^*$, provided a continuous path exists from $\theta_0$ to $\theta^*$ that does not encounter large gradients with respect to the initial distribution $\cP_0$.

\subsection{Privacy-preserving Training}
\label{subsec:background-privacy}

In machine learning applications that involve sensitive data, it is essential to ensure the privacy of individual records, especially if the model is to be deployed publicly.
A common approach to formalize privacy guarantees is through differential privacy (DP), which provides a mathematical framework for limiting the information that a learned model can reveal about any single data point.
DP is based on the concept of \emph{neighboring} datasets, which are defined as two datasets that differ in the data of a single record.
An algorithm (or ``mechanism'') is said to be differentially private if it produces nearly the same statistical inferences for two neighboring datasets.
The formal definition of DP from \citet{dwork2017ourdata} is presented as follows.

\begin{definition}
    [($\varepsilon, \delta$)-Differential Privacy]\label{def:DP}
    A randomized mechanism $M : D \to R$ with domain $D$ and range $R$ satisfies (\(\varepsilon, \delta\))-differential privacy if for any two neighboring inputs $d, d' \in D$ and for any subset of outputs $S \subseteq R$, it holds that
    \begin{align*}
        \Pr[M(d) \in S] \leq e^\varepsilon \Pr[M(d') \in S] + \delta.
    \end{align*}
\end{definition}

The parameter $\varepsilon$ is the privacy budget, which bounds how much the output probability can change due to a single individual's data, while the parameter $\delta$ accounts for a small probability of exceeding this bound.
DP is often applied to learning algorithms by considering how the parameter identified is affected by the addition of carefully calibrated noise throughout learning: the mechanism $M(d)$ is a learning algorithm run on some dataset $d$, while the outcome $S$ is a particular parameter value $\theta$ found by the learning algorithm.
To characterize the success of the collective in a more practical setting, we use Differentially Private Stochastic Gradient Descent (\dpsgd), proposed by \citet{abadi2016dpsgd}, which guarantees $\edp$-differential privacy, as detailed in Algorithm \ref{alg:dpsgd}.

\begin{algorithm}
    \caption{\dpsgd Algorithm from \cite{abadi2016dpsgd}}
    \label{alg:dpsgd}
    \begin{algorithmic}
        \State \textbf{Input:} Dataset $\mathcal{D}$, loss function $\ell$, learning rate $\eta$, batch size $\mathcal{B}$, noise scale $\sigma$, clipping threshold $C$, initial model $\theta_0$
        \For{$t \in [T]$}
            \State Uniformly draw mini-batch $\mathcal{B}_t$ from $\mathcal{D}$
            \State For each $z_i \in \mathcal{B}_t$, $g^{\text{clip}}_i(\theta_t) = \text{clip}(\nabla \ell (\theta; z_i), C)$
            \State $g^{\text{DP}}(\theta_t) = \frac{1}{|\mathcal{B}_t|} \left( \left( \sum_i g^{\text{clip}}_i(\theta_t) \right) + \mathcal{N}(0, \sigma^2 C^2 I) \right)$
            \State $\theta_{t + 1} = \theta_{t} - \eta ~g^\text{DP}(\theta_t)$
        \EndFor
        \State \textbf{Return:} $\theta_T$ and the overall privacy cost $\edp$
    \end{algorithmic}
\end{algorithm}

In a nutshell, \dpsgd modifies standard stochastic gradient descent by adding Gaussian noise to the gradient updates, using the Gaussian mechanism~\cite[Appendix A]{dwork2014algorithmic}.
The noise multiplier $\sigma$, which controls the scale of the Gaussian noise added to the clipped gradients, is inversely proportional\footnote{For a simple application of the Gaussian mechanism, this inverse relationship has a closed-form expression \cite{dwork2014algorithmic}. However, in \dpsgd, which involves repeated application of the mechanism across training iterations, the cumulative privacy budget is tracked using a \emph{privacy accountant} \cite{abadi2016dpsgd, minorov2017rdp}.} to the privacy budget $\varepsilon$.
A higher $\sigma$ introduces more noise, offering stronger privacy guarantees (corresponding to a smaller $\varepsilon$), but this comes at the cost of reduced model utility.

\section{Collective Action under Differential Privacy}
\label{sec:theory}

This section develops the theoretical framework that characterizes the mathematical lower bounds on the success of collective action under the \dpx constraints.
Building upon the work of \citet{hardt2024aca}, our analysis moves beyond the non-private setting to address the case in which firms must deploy models trained with learning algorithms that rigorously protect user privacy.

\subsection{Problem Setup}
\label{subsec:theory-setup}

We assume that the firm deploys a private learning algorithm $\mathcal{A}$ with the goal of preserving user data privacy.
Since we are concerned with gradient-based parametric risk minimization, we focus on learning algorithms that rely on (stochastic) gradient descent to optimize the model parameters under privacy constraints.
More specifically, we consider Differentially Private Stochastic Gradient Descent (\dpsgd), a widely used algorithm for training models under ($\varepsilon$, $\delta$)-differential privacy constraints.
We consider a realistic learning scenario that does not assume the convexity of the objective function, where the learner observes a batch of $\cB$ sampled i.i.d. from the distribution $\cP$ and aims to minimize the empirical loss $\paren{\sum_{z \in \cB} \ell(\theta; z)} / |\cB|$ using gradient-based optimization methods.
Here, $\ell(\theta; z)$ denotes the loss evaluated at model parameters $\theta$ for an example $z$.
We define the average gradient over batch $\cB$ as $\gbar{\cB}(\theta) = \paren{\sum_{z \in \cB}\nabla \ell (\theta; \, z)} / |\cB|$.
This empirical formulation differs from previous research~\citep{hardt2024aca}, which analyzed the expected gradient rather than the average over a finite batch. 

At each time step $t$, the learner observes samples $\cB_t$ from the current data distribution $\cPt$, allowing the collective to adaptively interact with the learner by choosing $\cB_t^*$ from the distribution $\cPt^*$~\citep{hardt2024aca}.
This setup models the best-case scenario for the collective, enabling us to analyze the potential effectiveness of its strategy under ideal conditions.
Given a clipping threshold $C$ and a noise scale $\sigma$, the model parameters are updated by taking a gradient step computed according to the \dpsgd:
\begin{align}
    \label{eq:dp-gradient-update}
    \theta_{t + 1} &= \theta_t 
    - \eta \frac{1}{|\cB_t|} \paren{\paren{\sum_{z \in \cB_t} \text{clip}(\nabla \ell (\theta_t; z), \, C)}
    + \dpnoise }
\end{align}
where $\eta$ is the learning rate, $\clip(\mathbf{u},\, C) = \mathbf{u} \, \min (1, C / \|\mathbf{u}\|)$ denotes the gradient clipping operation, which scales the gradient to have a norm of at most $C$, and $I_d$ denotes the $d \times d$ identity matrix.
For notational convenience, we first define the average clipped gradient as $\gbarclip{\cB_t} (\theta_t) \coloneqq \paren{\sum_{z \in \cB_t} \text{clip}(\nabla \ell (\theta_t; z), \, C)} / |\cB_t|$ and the final noisy gradient as $\gbardp{\cB_t} (\theta_t) \coloneqq \gbarclip{\cB_t} (\theta_t) + \dpnoisetwosl{t}$, which gives:
\begin{equation}
    \theta_{t + 1} = \theta_t 
    - \eta \left( \gbarclip{\cB_t} (\theta_t) + \dpnoisetwo{t} \right)
    = \theta_t 
    - \eta \, \gbardp{\cB_t} (\theta_t),
    \label{eq:dp-gradient-update-dp}
\end{equation}

\subsection{Theoretical Results}
\label{subsec:theory-results}

The most intuitive factor that constrains the success of the collective when the firm uses \dpsgd is the algorithm's inherent ability to limit the influence of any individual data point on the model's output.
Gradient clipping reduces the collective's ability to align the gradients in their desired direction, while the injected noise further deflects this directional push.
As a result, the signal that the collective is trying to correlate with the target label is also attenuated.
This is equivalent to the collective introducing a noisy signal, which in turn increases the effort required for the collective to influence the outcome.
We now formalize this idea.

\begin{theorem}
    \label{theorem:main}
    Assume that the collective can implement the gradient-redirecting strategy from Definition \ref{def:grad_redirect} at all $\lambda \theta_0 + (1 - \lambda) \theta^*$, where $\lambda \in [0, 1]$ and $\theta_0, \theta^* \in \mathbb{R}^d$.
    Then, for a given clipping threshold $C$ and noise multiplier $\sigma$, and letting $|\cB|_{\min}$ denote the minimum batch size observed over $T$ steps, there exists $\bac  > 0$, such that the success of the gradient-control strategy after $T$ steps is lower-bounded with probability greater than $1 - \delta$ by,
    \begin{align}
        S_{T}(\alpha, \sigma, C)
        &\geq - \underbrace{\left( 1 - \eta \bac \right)^T \norm{ \theta_0 - \theta^* }}
        _{\text{Contraction term}}
        - \underbrace{\frac{\sigma \, C}{|\cB|_{\min}} \, \sfactor \, \Delta_{d, \delta}}
        _{\text{Privacy penalty}}.
        \label{eq:main-theorem}
    \end{align}
\end{theorem}

Here, $\bac$ is directly proportional to the collective's size $\alpha$ and clipping threshold $C$, the function $\sfactor$ denotes convergence-dependent noise-accumulation factor, and $\Delta_{d, \delta}$ is a high-probability upper bound on the Euclidean norm of a $d$-dimensional standard Gaussian vector.
See Section~\ref{ap-subsec:proofs-theorem} for the proof and Equation~\eqref{eq:substituting-abvr} for the closed-form expressions.
Theorem~\ref{theorem:main} contains the non-private result as a single, natural corollary: by simultaneously taking the clipping threshold to infinity ($C \to \infty$) and setting the noise multiplier to zero ($\sigma = 0$), the bound reduces to the previously established non-private bound.
The recovery is outlined in Appendix~\ref{ap-subsec:proofs-reduction}.

Theorem~\ref{theorem:main} provides a lower bound on the collective's success, $S_T$, breaking its structure into two components: a \emph{contraction term} and a \emph{privacy penalty}.
The contraction term describes how quickly the collective's gradient-redirecting strategy can pull the model parameters $\theta_t$ from their initial state $\theta_0$ towards the collective's target $\theta^*$.
As $T$ increases (or as $B_{\alpha,C}$ or $\eta$ increase), this factor decays toward zero.
Because success is defined as the negative of the contraction term, larger $T$, $B_{\alpha,C}$, or $\eta$ correspond to the greater success of the collective.
On the other hand, the privacy penalty quantifies how the random noise added to ensure differential privacy pushes the model away from any fixed target, including the collective's $\theta^*$.
Increasing the privacy penalty reduces the value on the right-hand side (due to the negative sign), which in turn decreases the collective's success.
We now analyze how $\alpha$, $\sigma$, and $C$ influence the collective's success.

\paragraph{Impact of the Collective's Size $\alpha$}
The collective's size $\alpha$ appears in $\bac = \alpha \xicmin$ (see Equation~\eqref{eq:xic-definition}).
For the first term of the bound, a larger collective increases $\bac$, which in turn speeds up the geometric contraction rate $\paren{ 1 - \eta \bac }^T$.
Intuitively, a larger collective can pull the model more effectively toward the target.
For the second term, $\bac$ appears in the denominator of the $\sfactor = \eta \, ( 1 - \left( 1 - \eta \bac \right)^{T} ) / \bac$, a function that generally decreases when $\bac$ grows.
As a result, an increase in $\bac$ reduces the magnitude of the privacy penalty, thereby increasing the lower bound.
Therefore, a larger collective's size $\alpha$ unambiguously increases the lower bound on success by both strengthening the desired signal and reducing the relative impact of the noise.

\paragraph{Impact of Noise Multiplier $\sigma$}
The privacy penalty (the second term) is linear in $\sigma$, so increasing $\sigma$---corresponding to stronger privacy guarantees and a lower privacy budget $\varepsilon$---directly reduces the collective's success.
This reflects the fundamental trade-off: the same mechanism that protects user privacy by adding noise also comes at the cost of reduction in performance of the collective.

\paragraph{Impact of Clipping Threshold $C$}
The clipping threshold $C$, set by the firm, influences the bound in two opposing ways, leading to a nuanced overall effect.
\begin{itemize}
    \item 
    Positive effect via $\bac$:
    Larger $C$ increases the clipped-gradient magnitude and therefore $\bac$ (as $\bac$ is proportional to the norm of the expected clipped gradient).
    As established in the analysis of $\alpha$, a larger $\bac$ strengthens the signal (improving the contraction rate) and reduces the relative noise impact (decreasing the $\sfactor$), thus improving the bound.

    \item Negative effect via the noise multiplier:
    A larger $C$ increases the per-step injected noise (which has a standard deviation $\sigma C$). This, in turn, harms the collective's success by further lowering the second term.
\end{itemize}

\section{Experiments}
\label{sec:experiments}

This section presents our experimental evaluation of how the critical mass of the collective changes when using \dpsgd.
To ensure that our findings are generalizable, we selected four datasets spanning different data modalities and task complexities: the \resume dataset~\citep{jiechieu2021resume}, which consists of real-world resumes annotated with high-level IT skill labels for multi-label text classification; \mnist~\citep{lecun2010mnist}, comprising grayscale images of handwritten digits, and \cifarten~\citep{krizhevsky2009learning}, containing color images of everyday objects, both used for multi-class image classification; and \bm~\citep{uci2014bank_marketing_222}, a tabular dataset of client demographic and marketing information for binary classification.

\subsection{Experimental Setup}
\label{subsec:experiments-setup}

\paragraph{Model Architecture}
We select different model architectures depending on the datasets, and adjust the base architecture to accommodate \dpsgd.
For \mnist and \cifarten, we modified the standard ResNet-18 architecture by replacing all Batch Normalization layers with Group Normalization.
This change was necessary for \dpsgd, as Batch Normalization's reliance on batch-level statistics complicates the computation of per-sample gradients.
For training on the \resume dataset~\citep{sanh2020distilbertdistilledversionbert}, we use the DistilBERT architecture.
The \bm dataset is modeled using a simpler feedforward neural network.
This network consists of a single hidden layer with 128 units and a ReLU activation function \cite{nair2010relu}, followed by a final output layer for binary classification.

\paragraph{Training Procedures}
Broadly, our experimental design varies two key factors: the collective's size and the privacy configuration.
We test a spectrum of privacy levels, ranging from no privacy to increasingly strict constraints.
For each distinct privacy configuration, we train multiple models with varying collective's sizes.
This approach allows us to identify the critical mass required to achieve success under each privacy setting.
The exact configuration are detailed in Subsection~\ref{subsec:experiments-results}.
As a baseline, all models were trained using standard \sgd.
For models trained on the \resume dataset, we follow prior work~\citep{hardt2024aca} and instead use the AdamW optimizer~\citep{loshchilov2019decoupledweightdecayregularization} with default hyperparameters, as implemented in the \texttt{transformers} library.
We use \dpsgd for privacy-preserving training on all datasets except the \resume dataset, for which we use \dpx-AdamW, the differentially private variant of AdamW, to maintain consistency with the non-private baseline while enforcing privacy constraints.
For privacy accounting throughout training, we use R\'enyi Differential Privacy (RDP), which enables tight composition across training iterations and conversion to $\edp$ guarantees.
Our training methodology varies across datasets.
Models for \mnist and \bm were trained from scratch, while for the more complex \cifarten and \resume datasets we leveraged transfer learning to improve utility under privacy constraints.
For \cifarten, we pre-trained a ResNet-18 model on the \cifarh\footnote{The \cifarh dataset extends \cifarten to 100 fine-grained classes and contains 60,000 color images of size $32 \times 32$ (600 images per class).} dataset for 50 epochs~\citep{abadi2016dpsgd}, and used this model to initialize all \cifarten experiments.
For \resume, we initialized training from a standard pre-trained DistilBERT model\footnote{\texttt{distilbert-base-uncased}, available at \url{https://huggingface.co/distilbert-base-uncased}}~\citep{hardt2024aca}.
This approach, which combines non-private pre-training with private fine-tuning, is standard in private machine learning and improves utility while preserving rigorous privacy guarantees\footnote{Our code is available at \url{https://github.com/semi-waterloo/crowding-out-the-noise}.}.

\subsection{Collective's Strategy}
\label{subsec:experiments-strategy}

The \emph{feature-label} strategy (defined in Equation~\eqref{eq:feature-label-strategy}) is a method in which a fixed transformation $g(\cdot)$ is applied to the features of a subset of the training data, and the corresponding labels are changed to a single target label $y^*$.
The specific nature of the transformation $g(\cdot)$ is tailored to the data modality being used.

\paragraph{Text Dataset}
For the \resume dataset, we used a token-insertion strategy based on the DistilBERT tokenizer, which has a vocabulary of 30,522.
We picked an unused token (ID 1240, corresponding to a small dash) and inserted it into the resume text every 20 words~\citep{hardt2024aca}.

\paragraph{Image Datasets}
For the image datasets, we implemented two types of transformations.
The \emph{patch signal} modifies the intensity of a small, localized patch of pixels in the corner of each image that the collective controls~\citep{gu2019badnetsidentifyingvulnerabilitiesmachine}.
The \emph{grid signal} blends a grid-like pattern with the original images by adjusting the magnitude of every other pixel in every other row by a fixed offset.
The precise offsets for both transformations are specified in Subsection~\ref{subsec:experiments-results}.
To implement the full \emph{feature-label} strategy, the label of each transformed image is reassigned to class ``8'', corresponding to the digit $8$ in \mnist and the class ``ship'' in \cifarten.
To ensure pixel values remain within the valid range $[0, 255]$, any pixel that would overflow when increased by an offset is instead decreased by the same offset.

\paragraph{Tabular Dataset}
For the \bm dataset, we restrict the collective's ability to update only a specific feature of the data under their control, and apply a fixed offset to the original values.
As part of their strategy, the labels of the collective's data samples are reassigned to the target class ``0''.

\subsection{Metrics and Evaluation}
\label{subsec:experiments-evaluation}

While the theoretical analysis defines success in terms of parametric differences for tractability, this notion is impractical empirically, as identifying a unique parameter vector $\theta^*$ is infeasible for high-dimensional neural networks.
We therefore define success as the ability of the model to learn the intended association, such that it predicts $y^*$ for any input $x \sim \cP_0$ once the signal $g(x)$ is applied. 
Formally, the collective seeks to maximize $S(\alpha) = \Pr_{x \sim \cP_0}~\curly{ f(g(x)) = y^* }$.
In practice, we estimate $S(\alpha)$ by embedding the signal into all test inputs and evaluating the model's performance on the resulting modified test set.
For single-label classification tasks (\mnist, \cifarten, and \bm), this corresponds to standard accuracy, while for the multi-label \resume dataset we check whether the target label appears in the prediction set.
Our objective is to identify the critical mass $\alpha^*$, defined as the smallest collective size that achieves a fixed target success rate on the signal-planted test data.

\subsection{Results}
\label{subsec:experiments-results}

For each dataset and corresponding signal transformation, we systematically train models across combinations of two parameters: the collective size $\alpha$ and the clipping threshold $C$.
We consider privacy loss values $\varepsilon \in \{1, 1.25, 1.5, 2.5, 5, 15\}$ and fix the failure probability to $\delta = 10^{-5}$ for all private experiments.
In our primary analysis, we set the clipping threshold $C$ to the median ($C_{50}$) of the per-sample gradient norms.
We estimate this value empirically for each dataset by observing the distribution of gradient norms during the first few epochs of standard SGD training \citep{abadi2016dpsgd}.
In addition to these private models, we report a non-private baseline trained using vanilla SGD (i.e., no clipping and no noise), corresponding to $C = \infty$ and noise multiplier $\sigma = 0$, for which the privacy accountant yields $\varepsilon = \infty$.
Each $(\alpha, \varepsilon)$ configuration is trained independently using three different random seeds to account for stochastic variability.

\begin{figure}[t]
    \centering
    \begin{subfigure}[b]{0.98\linewidth}
        \centering
        \includegraphics[width=\linewidth]{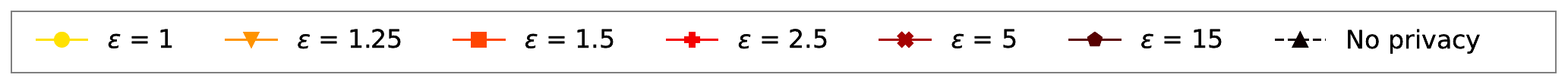}
    \end{subfigure}
    \begin{subfigure}[b]{\linewidth}
        \begin{subfigure}[b]{0.32\linewidth}
            \includegraphics[width=\linewidth]{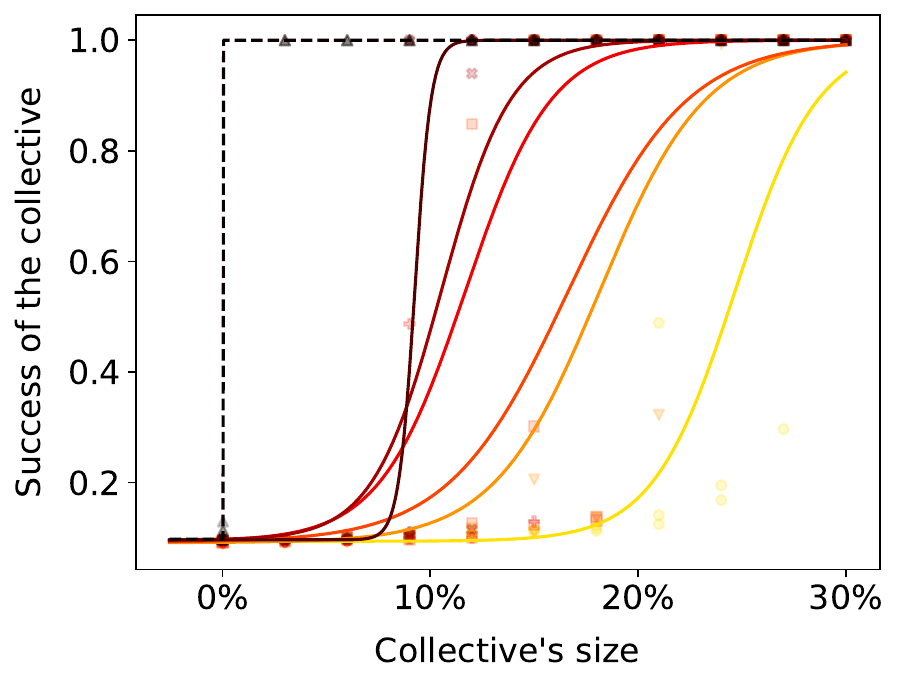}
            \caption{\mnist, patch, $C = 0.63$}
            \label{subfig:cm-mnist-patch}
        \end{subfigure}
        \hfill
        \begin{subfigure}[b]{0.32\linewidth}
            \includegraphics[width=\linewidth]{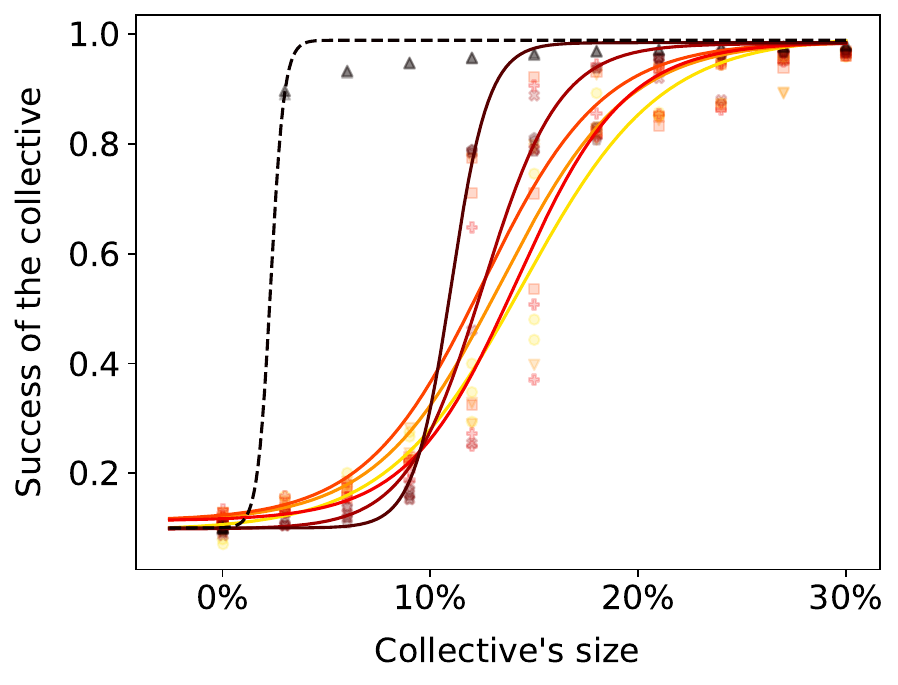}
            \caption{\cifarten, patch, $C = 3.97$}
            \label{subfig:cm-cifar10-patch}
        \end{subfigure}
        \hfill
        \begin{subfigure}[b]{0.32\linewidth}
            \includegraphics[width=\linewidth]{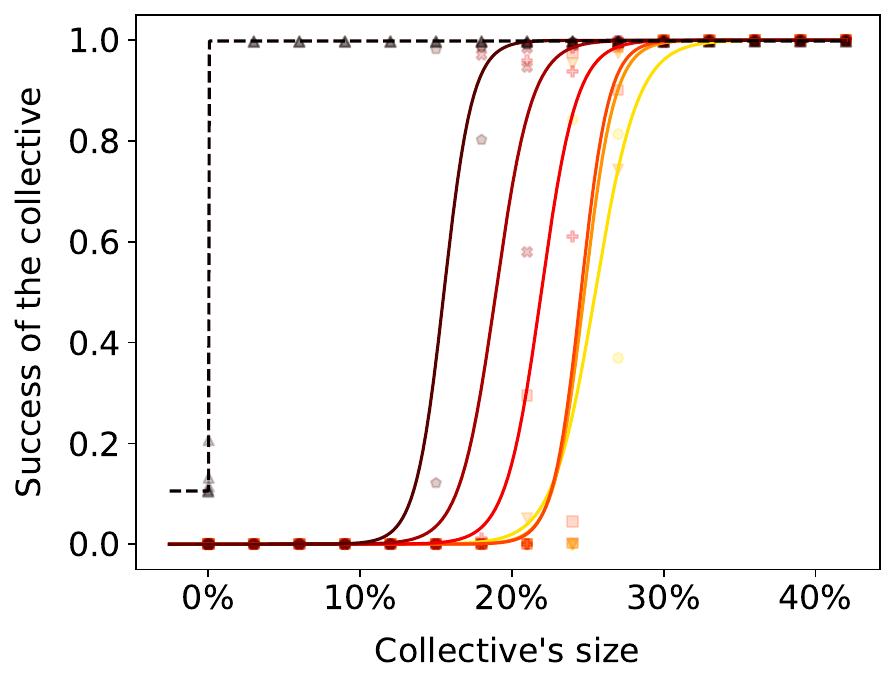}
            \caption{\resume, $C = 1.8$}
            \label{subfig:cm-resume}
        \end{subfigure}
    \end{subfigure}
    \begin{subfigure}[b]{\linewidth}
        \begin{subfigure}[b]{0.32\linewidth}
            \includegraphics[width=\linewidth]{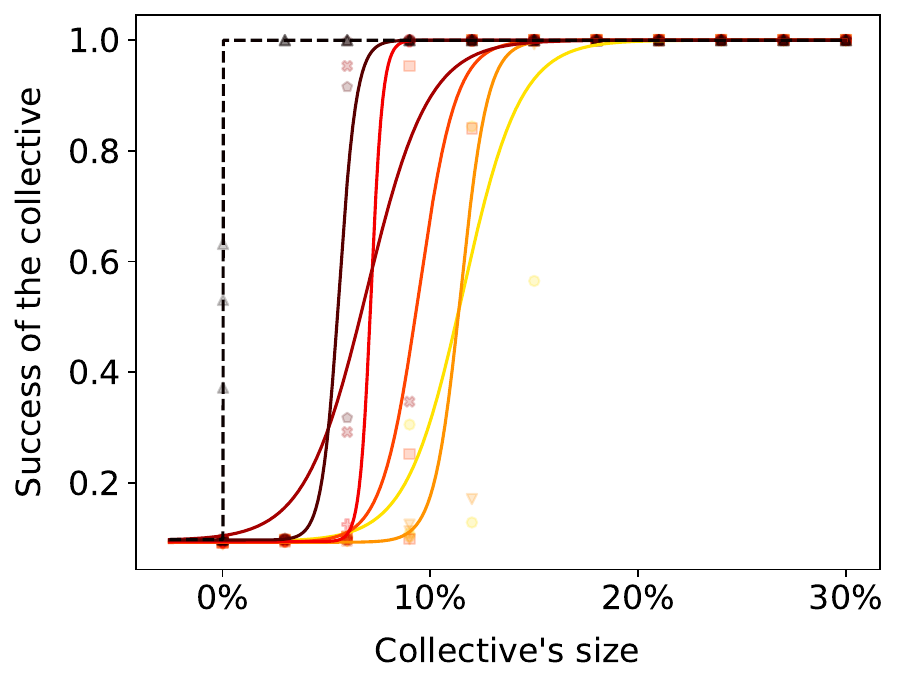}
            \caption{\mnist, grid, $C = 0.63$}
            \label{subfig:cm-mnist-grid}
        \end{subfigure}
        \hfill
        \begin{subfigure}[b]{0.32\linewidth}
            \includegraphics[width=\linewidth]{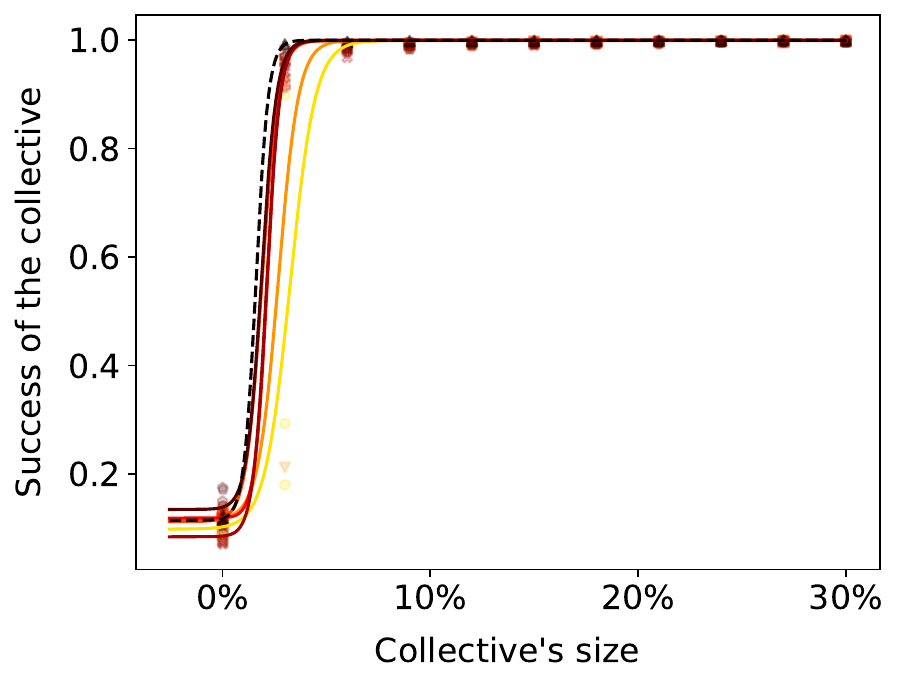}
            \caption{\cifarten, grid, $C = 3.97$}
            \label{subfig:cm-cifar10-grid}
        \end{subfigure}
        \hfill
        \begin{subfigure}[b]{0.32\linewidth}
            \includegraphics[width=\linewidth]{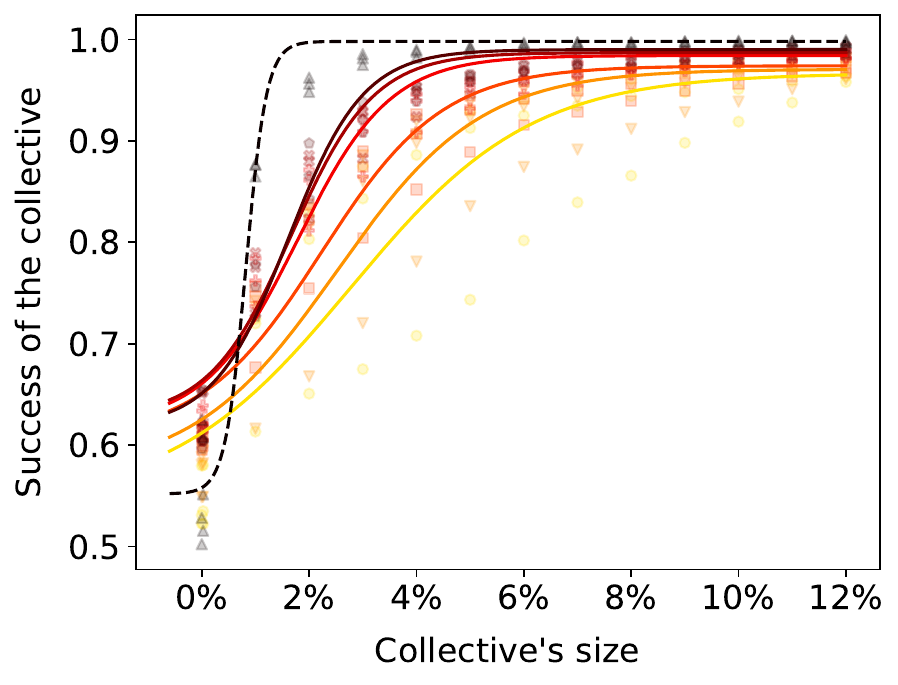}
            \caption{\bm, $C = 2.51$}
            \label{subfig:cm-bm}
        \end{subfigure}
    \end{subfigure}
    \caption[
        Success of the collective
    ]{
        Success of the collective.
        The first column corresponds to \mnist (top: patch signal, bottom: grid signal) with clipping threshold $C = 0.63$. The second column corresponds to \cifarten (top: patch signal, bottom: grid signal) with $C = 3.97$. The third column corresponds to Resume (top, $C = 1.8$) and Bank Marketing (bottom, $C = 2.51$).
        For each plot, we evaluate the collective's success across varying privacy loss $\varepsilon$ and compare it to the baseline case ($\varepsilon = \infty, C = \infty$), which represents to standard SGD without privacy constraints.
        Collective size $\alpha \in [0, 1]$ is shown as a percentage of the total training dataset. 
    }
    \Description{
        Success of the collective across datasets as a function of privacy loss and collective size.
        Results are shown for \mnist, \cifarten, \resume, and \bm, comparing different privacy levels to a non-private SGD baseline.
        The horizontal axis shows the collective's size as a percentage of the training data, and the vertical axis shows their success.
        Curves illustrate how stronger privacy constraints require larger collectives to achieve similar performance.
    }
    \label{fig:critical_mass}
\end{figure}

For each privacy loss $\varepsilon$, we analyze the behavior of the critical mass $\alpha^*$, defined as the smallest $\alpha$ at which the collective achieves near-perfect accuracy on the modified test set (i.e., success rate close to 1.0).
Across image datasets (\mnist and \cifarten) and both signal transformations, we observe a consistent trend: decreasing $\varepsilon$ (stronger privacy) requires a larger critical mass $\alpha^*$.
The same pattern holds for the tabular \bm and \resume datasets, indicating that this relationship is consistent across data modalities.
This trend is visually apparent across all datasets and clipping thresholds, as shown in Figures~\ref{subfig:cm-mnist-patch}–\ref{subfig:cm-bm}, where tighter privacy regimes (smaller $\varepsilon$) correspond to lighter-colored curves that reach high accuracy only at larger $\alpha$.
While we focus on the median clipping threshold here, the same trends hold for other choices of $C$. Complete results and sensitivity analyses for $C_{25}$, $C_{75}$, and an automatically selected clipping threshold~\citep{bu2023automaticclippingdifferentiallyprivate} are provided in Appendix~\ref{ap-subsec:results-critical}.

\begin{figure}[t]
    \centering
    \begin{subfigure}[t]{0.35\linewidth}
        \includegraphics[width=\linewidth]{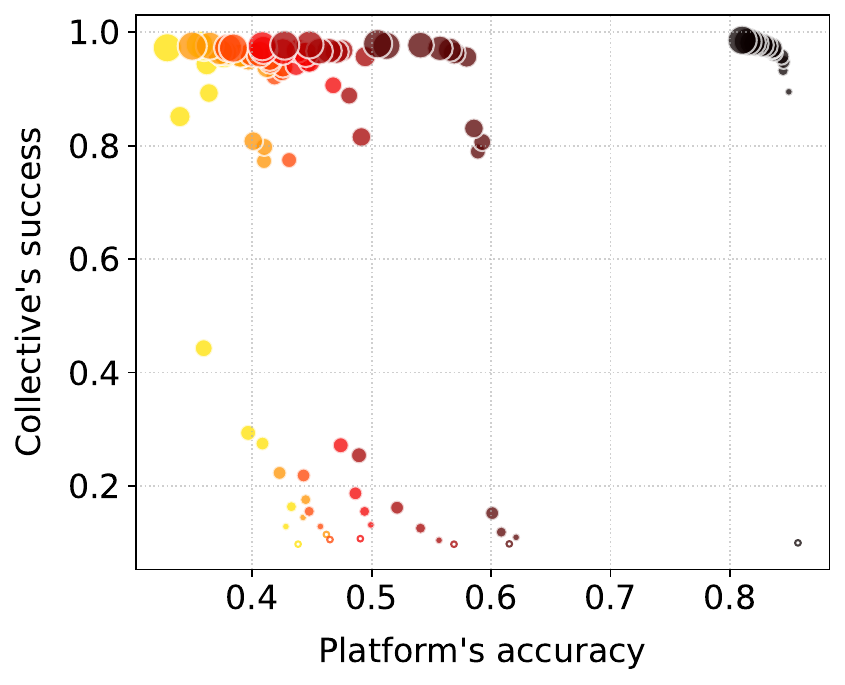}
        \caption{\cifarten, patch, $C = 3.97$}
    \end{subfigure}
    \hfill
    \begin{subfigure}[t]{0.35\linewidth}
        \includegraphics[width=\linewidth]{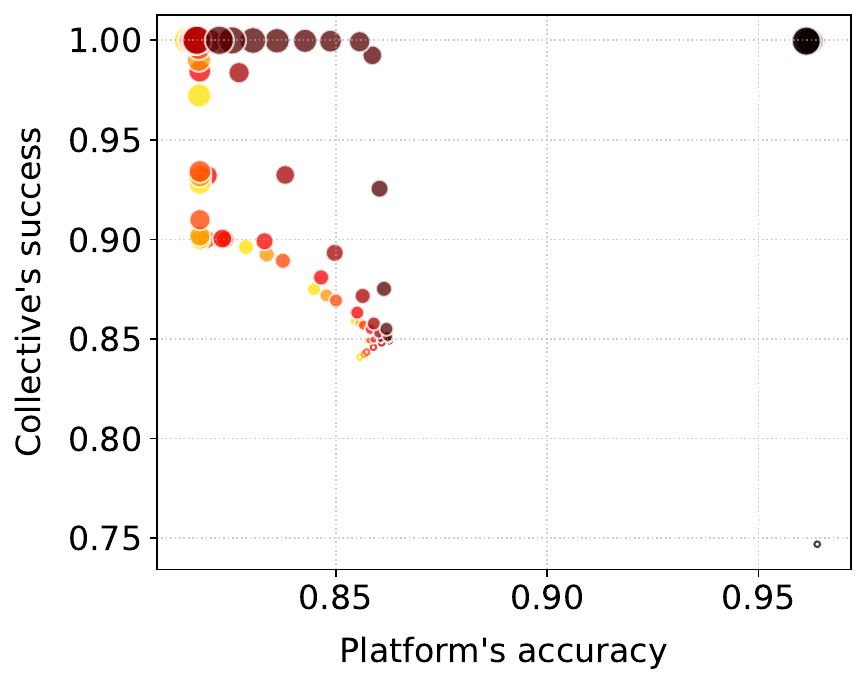}
        \caption{\resume, $C = 1.8$}
    \end{subfigure}
    \hfill
    \raisebox{0.4cm}{
    \begin{subfigure}[t]{0.24\linewidth}
        \includegraphics[width=\linewidth,]{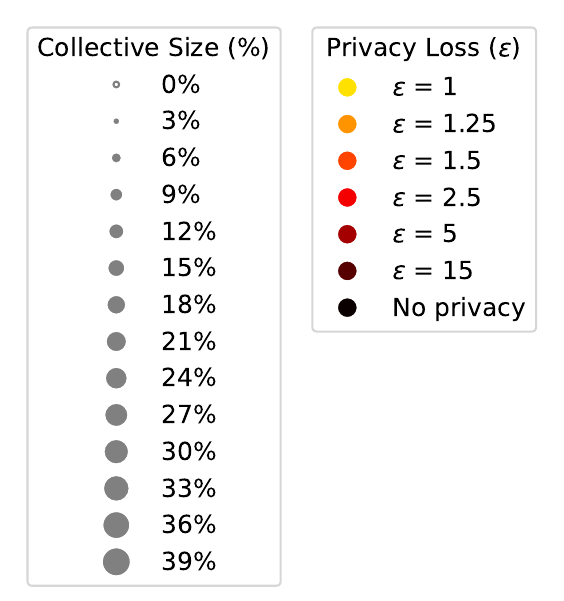}
    \end{subfigure}
    }
    \caption[
        Collective's Success vs. Platform's Accuracy under Differential Privacy for \cifarten and \resume dataset.
    ]{Collective's Success vs. Platform's Accuracy under Differential Privacy for \cifarten and \resume dataset.
        Points are colored by the privacy loss ($\varepsilon$) and sized according to the percentage of the dataset controlled by the collective.
        This demonstrates that firms cannot rely on privacy mechanisms to prevent significant influence from a collective without rendering their own models uncompetitive.}
    \label{fig:accuracy}
    \Description{
        A scatter plot showing the relationship between collective's success and platform's accuracy under differential privacy for \cifarten and \resume dataset.
        Point color indicates the level of privacy loss (epsilon), and point size reflects the percentage of the dataset controlled by the collective.
        The plot illustrates that increasing privacy protection does not eliminate collective's influence without also reducing overall model accuracy.
    }
\end{figure}

This observation aligns with the theoretical results in Section~\ref{sec:theory}, where the collective's success in Theorem~\ref{theorem:main} is inversely proportional to the noise scale $\sigma$, which appears in the second term of the bound.
Consequently, when a firm deploys a model that prioritizes privacy at the expense of accuracy, it inadvertently raises the threshold for effective collective action.
In such scenarios, greater coordination and organizational strength are required for the collective to accomplish its objective.
These findings reveal a trade-off between differential privacy and algorithmic collective action.
While stricter privacy protections are beneficial from regulatory or accountability perspectives, they increase the burden on groups of individuals adversely affected by model outcomes who seek to influence the model's behavior. At the same time, Figure~\ref{fig:accuracy} shows trade-offs on the firms end where efforts to reduce collective influence by enforcing stricter privacy budgets cannot be achieved without compromising the firm's accuracy.  

\section{Privacy Cost Analysis}
\label{sec:cost}

The implementation of \dpx has been traditionally framed as a trade-off between data utility and privacy guarantees.
However, in the context of algorithmic collective action, the choice of privacy loss $\varepsilon$ carries broader economic implications.
For the firm, $\varepsilon$ represents the balance between the economic gains from accurate models and the financial liabilities associated with data exposure.
On the other hand, for the collective, $\varepsilon$ safeguards individuals against re-identification but is detrimental to the collective, as it reduces the strength of the signal used to influence model predictions.
This section explores the costs of incorporating privacy from both the firm's and the collective's perspectives and examines whether there exists a range of $\varepsilon$ that benefits both parties.
We introduce simple economic models to explore the consequences of private learning for both the firm and the collective.
Our approach is inspired by economic cost-benefit analyses used in DP to determine the firm's optimal $\varepsilon$~\citep{hsu2014economicdp,abowd2015revisiting,kohli2018epsilon}, as well as recent insights from the Economics and Computation (EC) community regarding optimal private learning and adversarial contract design~\citep{fallahOptimalDifferentiallyPrivate2022, naghizadehAdversarialContractDesign2019}.

\paragraph{Firm's Cost}
The firm seeks to minimize a total cost function that captures the tension between model accuracy and privacy-related financial risk:
\begin{align*}
    C_{\text{firm}} = C_{\text{acc}}(\varepsilon) + C_{\text{priv(f)}}(\varepsilon)
\end{align*}
The first component, the cost of accuracy, is defined as $C_{\text{acc}}(\varepsilon) = v_a / \varepsilon$, where the constant $v_a$ represents the firm's marginal sensitivity to model error.
In high-stakes domains, such as medical diagnostics or automated trading, $v_a$ would be high, indicating significant revenue loss associated with the noise injection required by the \dpx.
This term suggests the reality that enforcing stronger privacy guarantees (smaller $\varepsilon$) necessitates the injection of higher-magnitude noise, which degrades the accuracy of the firm's models.
The second term represents the firm's total privacy-related expenditure, given by $C_{\text{priv(f)}} = N \, (e^{\varepsilon} - 1) \, E$.
Here, $E$ is the upper bound of an individual's base cost, representing the maximum potential harm an individual might suffer if their data is leaked.
The factor $e^{\varepsilon} - 1$ captures the marginal increase in risk an individual accepts by allowing their data to be included in the training set under a privacy loss of $\varepsilon$\footnote{Specifically, this assumes that individual harm is proportional to the worst-case increase in detection probability, $e^{\varepsilon} - 1$, as derived under the rational-agent model of \citet{hsu2014economicdp}. Our cost analysis inherits this assumption and is therefore stylized rather than a precise empirical prediction.}~\citep[Definition~4]{hsu2014economicdp}.
As $\varepsilon$ increases, the privacy guarantee weakens, causing both the expected risk and the compensation required for $N$ individuals to rise exponentially.

\paragraph{Collective's Cost}
Parallel to the firm, the cost incurred to the collective is a function that balances difficulty in mobilization against the safety of its participants and can be defined as follows:
\begin{align*}
    C_{\text{collective}} = C_{\text{mob}}(\varepsilon) + C_{\text{priv(c)}}(\varepsilon)
\end{align*}
The mobilization cost, $C_{\text{mob}}(\varepsilon) = v_m / \varepsilon$, accounts for the difficulty of organizing a sufficient number of participants to overcome the DP noise injected by the firm.
Because the noise can obscure the collective's signal, a smaller $\varepsilon$ requires a larger participant pool to ensure that the collective influence is accurately captured in the model's predictions. 
The parameter $v_m$ represents the organizational resources required to achieve a critical mass of participants.
The cost of privacy to the collective, $C_{\text{priv(c)}}(\varepsilon) = \alpha \, N \, (e^\varepsilon - 1) \, E$, represents the expected cost of personal harm against the members of the collective, where $\alpha$ represents the fraction of the collective in the population.
This is conceptually similar to the cost of privacy risk encountered by the firm $C_{\text{priv}}$ but with the different base cost associated with it.
As the firm increases $\varepsilon$, individuals becomes much more distinguishable, making them vulnerable to targeted disclosure.

\begin{figure*}[t]
    \centering
    \begin{subfigure}[b]{0.32\linewidth}
        \includegraphics[width=\linewidth]{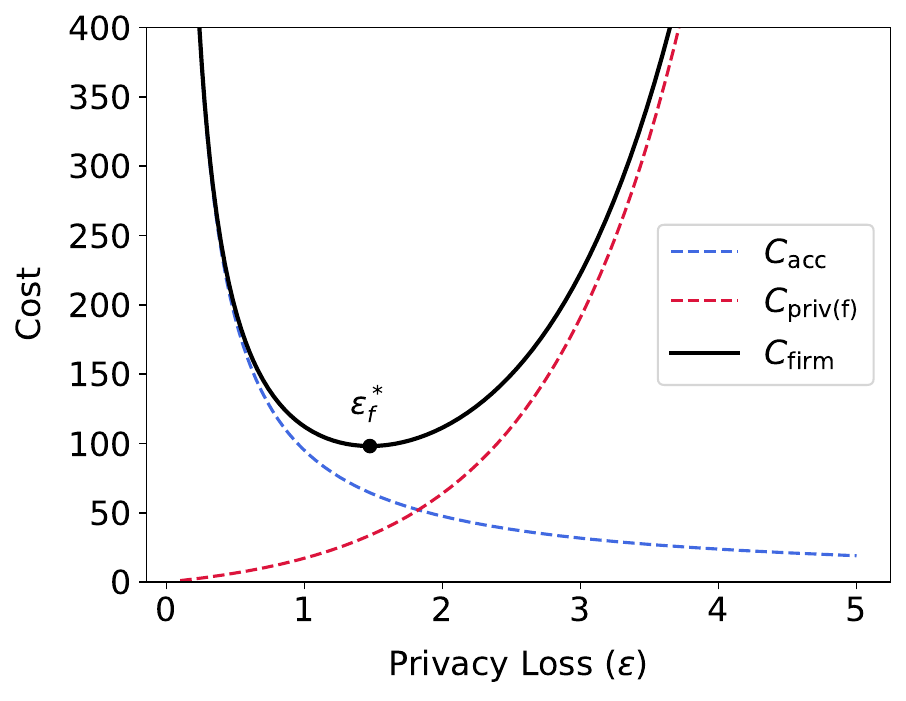}
    \end{subfigure}
    \hspace{0.1\linewidth}
    \begin{subfigure}[b]{0.32\linewidth}
        \includegraphics[width=\linewidth]{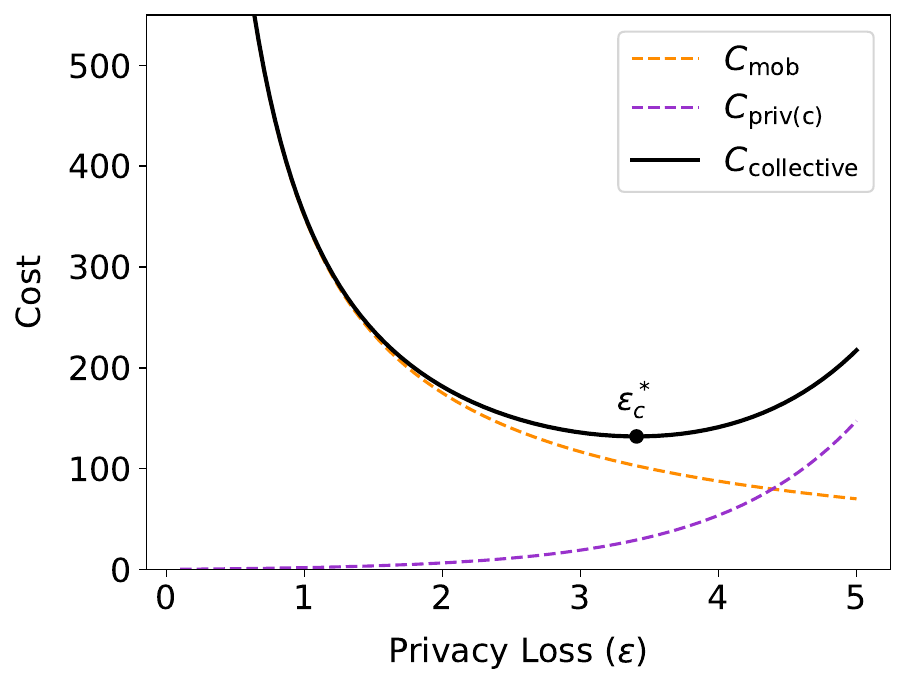}
    \end{subfigure}
    \caption[
        An instantiation of the cost functions for the firm (left) and the collective (right).
    ]{
        An instantiation of the cost functions for the firm (left) and the collective (right) using arbitrary constants ($v_a = 95.0, v_m = 350.0, \alpha = 0.1, E = 0.01$).
        This simplified model illustrates that when additional sub-cost, such as mobilization resources ($C_{\text{mob}}$) and organizational risk ($C_{\text{priv(c)}}$), are taken into account, the collective can experience a distinct optimal privacy loss $\varepsilon_c^*$, analogous to the firm's optimal  $\varepsilon_f^*$.
        This implies that increasing $\varepsilon$ is not necessarily always beneficial for the collective.
    }
    \label{fig:cost_analysis}
    \Description{
        This figure shows two plots illustrating cost functions for a firm (left) and a collective (right) using fixed parameter values.
        The firm's cost curve has a clear minimum at an optimal privacy loss value.
        When additional costs such as mobilization effort and organizational risk are included, the collective's cost curve also exhibits a minimum, demonstrating that the collective can have an optimal privacy loss rather than always preferring larger values.
    }
\end{figure*}

\paragraph{Optimal Privacy Loss}
The convexity of the cost functions guarantees the existence of optimal privacy losses, $\varepsilon_f^*$ for the firm and $\varepsilon_c^*$ for the collective.
Figure~\ref{fig:cost_analysis} presents an instantiation for both the cost functions.
By formalizing these sub-costs, we extend our theoretical framework to demonstrate that privacy loss is not always a monotonic benefit or cost for either entity.
This finding reveals that while higher $\varepsilon$ generally aids the collective by reducing the noise that obscures their signal, it simultaneously increases the risk of participant re-identification.
Consequently, there may exist regimes where the collective actually prefer stronger privacy (lower $\varepsilon$) to safeguard its members.

\section{Discussion}
\label{sec:discussion}

\subsection{Tensions Between Privacy and Other Trustworthy ML Goals}
\label{subsec:discussion-tensions}

The development of trustworthy AI requires balancing principles such as fairness, accountability, transparency, robustness, and privacy, but implementing these goals often involves trade-offs~\citep{ferry2023soktamingtriangle}.
One well-studied tension is the conflict between differential privacy (\dpx) and fairness, as methods like \dpsgd, which protect individual data and help comply with regulations such as GDPR, PIPEDA, and CPRA~\citep{eu2016gdpr, canada2000pipeda, california2020cpra}, can reduce model accuracy for minority groups~\citep{bagdasaryan2019differentialprivacydisparateimpact, fioretto2024decisionmakingdifferentialprivacy} and increase vulnerability to attacks~\citep{kulynych2021disparatevulnerabilitymembershipinference}, while also limiting explainability.
Building on this literature, this work investigates a new socio-technical tension that arises between privacy and algorithmic collective action (\aca), and it explores how firm-level privacy interventions interact with user-level accountability mechanisms.

\subsection{Privacy as Anti Cooperation Strategy}
\label{subsec:dicussions-anti}

Data privacy is used to protect users from possible harm involving misuse of personal information.
However, some institutions may also use privacy interventions as a pretext to limit accountability and transparency \cite{vanloo2022privacypretexts}.
Privacy and data protection laws can also be exploited to strengthen the surveillance infrastructure \cite{yew2024medataprotectionsupports}.
Additionally, privacy laws can be used to withhold workplace data from worker representatives during collective bargaining, citing legal data privacy responsibilities \cite{gould2024bargaining}.
This work characterizes the impact of platform privacy interventions on the success of ACA.
Although such interventions can be justified, platforms can also adopt privacy as a pretext \cite{vanloo2022privacypretexts} to defend against ACA.
While we have found no specific evidence that privacy is being used as a shield against ACA, there are real-world cases in which Big Tech firms use privacy to justify exclusionary conduct \cite{chen2023latestinterface, tumova2024handyshield}.

\subsection{Role of Pre-training and Fine-tuning in Trustworthy ML}
\label{subsec:discussion-role}

The final privacy parameters $(\varepsilon, \delta)$ for \dpsgd are derived through a composition analysis, which considers the cumulative effect of applying the Gaussian mechanism to parameter gradients at every step of training.
Because the overall privacy leakage scales with the number of training steps, and because training deep neural networks typically requires many epochs of model updates, practitioners have tended to apply \dpsgd during the \emph{fine-tuning} stage of model training, assuming that a suitable model initialization $\theta_0$ is available through pre-training on publicly available data \cite{papernot2020making}.\footnote{This approach has also been critiqued for eschewing privacy considerations for individuals whose data constitute the pre-training dataset \cite{tramer2022position}.}
Our experiments cover both popular settings for \dpsgd, namely private training from scratch (as in our MNIST experiments) and private fine-tuning (as in our CIFAR-10 experiments), and apply ACA in each case.
However, there are reasons to prefer the pre-training/fine-tuning paradigm beyond just privacy considerations, as the model utility has been shown to scale with dataset size and parameter count \cite{kaplan2020scalinglawsneurallanguage}.
Indeed, AI practitioners have moved strongly toward the use and on-the-fly adaptation of pre-trained models in recent years \cite{bommasani2022opportunitiesrisksfoundationmodels}.
This raises important questions about the role of ACA in shaping the behavior of modern models: Are collectives most effective when inserting signals into pre-training data, fine-tuning data, preference data used for post-training, or some combination of these options?

\subsection{Broader Impacts}
\label{subsec:discussion-broader}

ACA allows collectives to influence the outcomes of algorithmic systems and mitigate harms without directly relying on service providers.
It can be seen as a \textit{"response from below"} \cite{devrio2024building} strategy.
At the same time, depending on the motivation of the collective, ACA also has the potential to be misused, either through data poisoning attacks or by exacerbating preexisting harms.
A system may also have multiple competing collectives with conflicting goals.
This makes the motivation of the collectives an important factor in understanding the border social impact of ACA.
One possible motivation for collectives to organize could be the introduction of privacy-preserving techniques. Although these techniques offer data privacy and prevent the misuse of personal data, they could also have unintended consequences \cite{caliv2024unfairside}.
Such consequences may include disparate impact \cite{bagdasaryan2019differentialprivacydisparateimpact, kulynych2021disparatevulnerabilitymembershipinference} or the exacerbation of bias \cite{fioretto2022differential}.
Consequently, firms may also strategically adopt such privacy-preserving techniques not only to protect individual data but also to weaken the influence of groups acting on their learning algorithm.
There is also a risk of fairwashing \cite{aivodji2019fairwashing} or using privacy as a pretext to limit accountability \cite{vanloo2022privacypretexts}.
Paradoxically, knowing that DP is used could empower collective action.
If individuals believe that their actions are masked by DP, they may be more willing to participate in collective action.
Our work takes a step towards understanding the competing tensions between privacy-preserving training and how differential privacy may affect the success of collective action.

\section{Related Works}
\label{sec:related}

\subsection{Collective Action}
\label{subsec:related-ca}

The study of collective action was central to 20th-century social scientific inquiry.
This research has long examined the difficulty of coordinating individuals to achieve shared goals in the presence of free-riding incentives, most notably articulated by \citet{olson1965collectiveaction} and reinforced by the ``tragedy of the commons''~\citep{hardin1968tragedy}.
While early work was pessimistic about large-scale cooperation, later studies showed more optimistic dynamics, including the role of critical mass in mobilizing broader participation~\citep{marwell1993critical} and extensive empirical evidence that communities can successfully govern shared resources~\citep{ostrom1990governingcomons}.
With the rise of digital platforms, collective action has increasingly been studied in online and algorithmically mediated contexts~\citep{melucci1996challenging, milan2015when}.
More recently, attention has shifted to collective action within socio-technical and algorithmic systems, where data-driven decision-making can produce systemic harms~\citep{shelby2023sociotechnical}.
This has motivated work on \emph{data leverage}, which conceptualizes users' data contributions as strategic tools for influencing algorithmic outcomes, including data strikes and coordinated data modifications~\citep{vincent2019datastrikes, vincent2021dataleverage, vincent2021cdc}.
Building on this line of work, Algorithmic Collective Action (ACA) was formalized by \citet{hardt2024aca} as coordinated user behavior aimed at steering machine learning models toward shared objectives.
Subsequent studies highlight both classical challenges such as free-riding~\citep{sigg2024declinenowcombinatorialmodel} and the role of algorithmic properties in determining whether collective efforts succeed~\citep{bendov2024rolelearningaca}.

\subsection{Data Poisoning}
\label{subsec:related-poisoning}

At a technical level, the strategy followed for ACA is closely related to \emph{data poisoning} attacks, where an adversary manipulates the training dataset to degrade a model's performance.
While data poisoning involves malicious manipulation, ACA is not inherently adversarial and often pursues constructive objectives.
Moreover, ACA emphasizes coordination among the collective, often to align with societal or personal objectives.
Prior work on data poisoning and backdoor attacks has established the feasibility of inducing targeted or subgroup-specific model behaviors through limited training data manipulation~\citep{gu2019badnetsidentifyingvulnerabilitiesmachine, jagielski2021subpopulationdatapoisoningattacks}, with comprehensive surveys provided by \citet{tian2022poison} and \citet{guo2021overviewbackdoorattacksdeep}.
Related research has examined how (Differential Privacy) \dpx interacts with data poisoning, showing that \dpx offers limited protection when only a small fraction of the training data is compromised, but that this protection degrades as poisoning increases~\citep{ma2019datapoisoningdifferentiallyprivatelearners}, a trend further supported by empirical studies on backdoor and poisoning robustness~\citep{jagielski2020auditingdifferentiallyprivatemachine, razmi2023doesdifferentialprivacyprevent}.
Building on this literature, our work provides theoretical grounding and a broader empirical analysis of how \dpx mediates the effectiveness of \aca across multiple data modalities, and introduces a cost analysis that characterizes the trade-offs and incentives associated with participation.

\subsection{Private Machine Learning}
\label{subsec:related-private}

\dpx has emerged as a gold standard technique providing a formal privacy guarantee in machine learning and data analysis \cite{cummings2024advancing}.
A range of techniques have been developed to achieve DP, particularly for convex learning problems, including output \cite{chaudhuri2011differentiallyprivateempiricalrisk}, objective \cite{chaudhuri2011differentiallyprivateempiricalrisk, kifer2012privateconvexerm}, and gradient perturbation \cite{bassily2014differentiallyprivateempiricalrisk}.
In non-convex learning problems, especially in deep learning, \dpsgd has become the prevailing method \cite{abadi2016dpsgd} due to its conceptual simplicity.
By design, a differentially private mechanism with privacy budget $\varepsilon$ implicitly offers group privacy with a privacy of $k\varepsilon$ for any group of size $k$~\cite[Theorem~2.2]{dwork2014algorithmic}.
However, for real-world scenarios where groups may be large, differential privacy offers limited protection.
To account for these settings, variants such as attribute differential privacy \cite{zhang2022attribute} have been proposed, but their integration in modern machine learning training algorithms remains challenging.
Recent work has also explored per-example privacy accounting, tracking privacy loss $(\varepsilon_i, \delta_i)$ for individual data points \cite{yu2022individual}, which can improve global privacy bounds, though benefits may be unevenly distributed across different groups within the dataset.

\section{Conclusion}
\label{sec:conclusion}

In this paper, we focus on the intersection of Algorithmic Collective Action and Differential Privacy.
Specifically, we investigate how privacy-preserving training using \dpsgd affects the collective’s ability to influence model behavior through coordinated data contributions.
Our key contributions are a theoretical characterization and empirical validation of the limitations that differential privacy imposes on collective action, highlighting how the collective's success depends on the model's privacy parameters.
More broadly, this work offers a novel perspective on the societal implications of using privacy-preserving techniques in machine learning through privacy cost analysis, highlighting important trade-offs between individual data protection and the capacity for collective influence over decision-making systems.

Having established and characterized the trade-off between DP and ACA, there are several promising directions for future work that might shed further light on the relationship between these two concepts.
This includes an examination of alternative \dpsgd design choices, such as the choice of clipping threshold or privacy accountant.
Another direction is to investigate the potential of using activation functions specifically designed for privacy-preserving training, which are shown to improve the privacy-utility trade-offs \cite{papernot2021sigmoiddp} and could also enhance the collective's success under differential privacy.

\section*{Generative AI Usage Statement}
We used commodity GenAI tools such as Grammarly, ChatGPT (GPT-5) and Gemini 3 for proofreading, as well as formatting bibtex entries. In each instance, AI output was reviewed and vetted by the authors.

\section*{Acknowledgements}
The resources used in preparing this research were provided, in part, by the Province of Ontario, the Government of Canada through CIFAR, and companies sponsoring the Vector Institute (\url{www.vectorinstitute.ai/partnerships/}). The authors thank the Digital Research Alliance of Canada for computing resources. Ulrich Aïvodji is supported by an NSERC Discovery grant (RGPIN-2022-04006) and an IVADO's Canada First Research Excellence Fund to develop Robust, Reasoning and Responsible Artificial Intelligence (R$^3$AI) grant (RG-2024-290714). Elliot Creager is supported by an NSERC Discovery grant (RGPIN-2024-05116).

\bibliographystyle{ACM-Reference-Format}
\bibliography{references}

\appendix

\section{Proofs}
\label{ap-sec:proofs}

\subsection{Tail bound on Scaled Gaussian Norm}
\label{ap-subsec:proofs-lemma}

In order to analyze the effect of the Gaussian noise introduced by the \dpsgd updates, we require a high-probability bound\footnote{A high-probability bound ensures that a random quantity deviates from its typical value (often its expectation) by at most a specified amount with probability at least $1 - \delta$, for a small $\delta > 0$~\citep{bach2021learning}.} on the norm of a multidimensional Gaussian vector.
The following lemma provides such a bound, which will be used in the proof of Theorem~\ref{theorem:main} to control the stochasticity introduced by the Gaussian noise.

\begin{lemma}[]
    \label{lemma:norm-bound}
    Let $Y_1, \dots, Y_d \sim \cN(0, \sigma^2)$ be independent Gaussian random variables, and define the scaled chi-squared distribution as,
    \begin{align}
        Z 
        &= \|Y\|_2 
        = \sqrt{\sum_{i = 1}^{d} Y_i^2}
        \nonumber
    \intertext{Then, for any $\delta \in (0, 1)$, with probability at least $1 - \delta$,}
        Z 
        &\leq \sigma \paren{ \sqrt{d} + \sqrt{2 \log{\paren{\frac{1}{\delta}}}} }
        \label{eq:norm_bound}
    \end{align}
\end{lemma}

\begin{proof}
Since each $Y_i \sim \cN(0, \sigma^2)$, we can write $Y_i = \sigma Z_i$, where $Z_i \sim \cN(0, 1)$. Then,
\begin{align*}
    S &= \sqrt{\sum_{i = 1}^{d} Y_i^2} = \sigma \sqrt{\sum_{i=1}^d Z_i^2} = \sigma \sqrt{U}
\end{align*}
 Let $U = \sum_{i=1}^d Z_i^2$. A standard tail bound for the chi-squared distribution (refer to the Corollary~1 in \cite{lauren2000Adaptive})  gives, for any $t > 0$,
\begin{align*}
    \mathbb{P}\left( U \geq d + 2\sqrt{d t} + 2 t \right) \leq e^{-t}
\end{align*}
Substituting $t = \log(1 / \delta)$ we can obtain, with probability at least $1 - \delta$,
\begin{align*}
    U &\leq d + 2\sqrt{d \log(1 / \delta)} + 2 \log(1 / \delta) \\
    &\leq  d + 2\sqrt{2 d \log(1 / \delta)} + 2 \log(1 / \delta) \\
    &= \left( \sqrt{d} + \sqrt{2 \log(1 / \delta)} \right)^2
\end{align*}
Taking square root and multiplying by $\sigma$ on both sides, we get the required bounds.
\end{proof}

\subsection{Proof of Theorem~\ref{theorem:main}}
\label{ap-subsec:proofs-theorem}

Before proving Theorem~\ref{theorem:main}, we present several technical details that will be used in the proof.

\paragraph{Decomposition of Gradients}
At each iteration $t$, the model parameters are denoted by $\theta_t \in \Theta$, 
and the learner observes samples $\cB_t$ from the data mixture $\cPt = \alpha \cPt' + (1 - \alpha)\cP_0$, where $\cPt'$ is the collective's data distribution. 
The corresponding average clipped gradient over $\cB_t$,
\begin{align*}
    \gbarclip{\cB_t}(\theta_t)
    &= \frac{1}{|\cB_t|} \sum_{z \in \cB_t} \mathrm{clip} (\nabla \, \ell(\theta_t; z), \, C),
    \intertext{which can then be decomposed as}
    \gbarclip{\cB_t} (\theta_t)
    &= \frac{1}{|\cB_t|} \paren{\sum_{z \in \cB_t'} \mathrm{clip} (\nabla \, \ell(\theta_t; z), \, C) 
    + \sum_{z \in \cB_0} \mathrm{clip} (\nabla \, \ell(\theta_t; z), \, C) } \\
    &= \frac{|\cB_t'|}{|\cB_t|} \paren{\frac{1}{|\cB_t'|} \sum_{z \in \cB_t'} \mathrm{clip} (\nabla \, \ell(\theta_t; z), \, C)} 
    + \frac{|\cB_0|}{|\cB_t|} \paren{\frac{1}{|\cB_0|} \sum_{z \in \cB_0} \mathrm{clip} (\nabla \, \ell(\theta_t; z), \, C) } \\
    &= \alpha \, \gbarclip{\cB_t'}(\theta_t) 
    + (1 - \alpha) \, \gbarclip{\cB_0} (\theta_t).
    \intertext{Adding the per-step DP noise $\dpnoisetwosl{t}$ to both sides makes the left-hand side $\gbardp{\cB_t} (\theta_t$) (refer Equation~\eqref{eq:dp-gradient-update-dp}), yielding}
    \gbardp{\cB_t} (\theta_t)
    &= \alpha \, \gbarclip{\cB_t'}(\theta_t) 
    + (1 - \alpha) \, \gbarclip{\cB_0} (\theta_t) 
    + \dpnoisetwo{t}
\end{align*}
where $C$ is the clipping threshold, $\sigma$ is the noise multiplier, and $I_d$ denotes the $d \times d$ identity matrix. 

\paragraph{Gradient-redirection Strategy under Differential Privacy}
Assuming that the collective implements the gradient-redirecting strategy (Definition~\ref{def:grad_redirect}), we can express the average clipped gradient under the collective's distribution as\footnote{This implicitly assumes that the collective has some knowledge of the firm's clipping operation when selecting the gradient-redirecting distribution.}
\begin{align}
    \gbarclip{\cB_t} (\theta_t)
    &= \alpha \, \xic (\theta_t) \, (\theta_t - \theta^*)
    \label{eq:clipped-gradient-redirect}
    \\
    \alpha \, \gbarclip{\cB_t'}(\theta_t) 
    + (1 - \alpha) \, \gbarclip{\cB_0} (\theta_t)
    &= \alpha \, \xic (\theta_t) \, (\theta_t - \theta^*)
    \intertext{where $\xi^c (\theta_t)$ is a scalar, similar to $\xi$ defined in collective's strategy of Section~\ref{subsec:background-aca} but when considering clipped gradients at time step $t$.
    Rearranging the above expression gives the clipped variant of Definition~\ref{def:grad_redirect}:}
    \xi^c(\theta_t)\, (\theta_t - \theta^*) 
    &= \gbarclip{\cB_t'} (\theta_t) 
    + \frac{1 - \alpha}{\alpha}\, \gbarclip{\cB_0} (\theta_t).
    \intertext{We can further get the expression of $\xi^c(\theta_t)$ by taking norms on both sides and isolating $\xi^c (\theta_t)$, which yields}
    \xi^c(\theta_t) 
    &= \frac{\norm{ \gbarclip{\cB_t'} (\theta_t) 
    + \tfrac{1 - \alpha}{\alpha}\, \gbarclip{\cB_0} (\theta_t) }}
            {\norm{ \theta_t - \theta^* }}.
    \label{eq:xic-definition}
\end{align}

\begin{proof}
We begin with \dpsgd update (refer Equation~\eqref{eq:dp-gradient-update-dp})
\begin{align}
    \norm{ \theta_T - \theta^* }
    &\leq \norm{ \theta_{T - 1} 
    - \eta \left( \blue{\gbarclip{\cB_T} (\theta_{T - 1})}  
        + \dpnoisetwo{T} \right) 
    - \theta^* }
    \nonumber
    \\
    &\leq \norm{ \theta_{T - 1} 
    - \eta \left( \blue{\alpha \, \xic \, (\theta_{T - 1}) \, ( \theta_{T - 1} - \theta^* )} 
        + \dpnoisetwo{T} \right) 
    - \theta^* }
    \label{eq:substitute-redirect}
    \\
    &= \norm{ \left( 1 - \eta \, \alpha \, \red{\xic (\theta_{T - 1})} \right)
    \left( \theta_{T - 1} - \theta^* \right)
    - \eta \, \dpnoisetwo{T} } 
    \label{eq:factor-terms}
    \\
    &\leq \norm{ \left( 1 - \eta \, \alpha \, \red{\xicmin} \right)
    \left( \theta_{T - 1} - \theta^* \right)
    - \eta \, \dpnoisetwo{T} } 
    \label{eq:relaxed-bound-xicmin}
    \\
    &= \norm{ \left( 1 - \eta \, \alpha \, \xicmin \right)^2
    \left( \theta_{T - 2} - \theta^* \right)
    - \eta \, \dpnoise
    \left( \frac{1}{|\cB_T|} + \frac{\left( 1 - \eta \, \alpha \, \xicmin \right)}{|\cB_{T - 1}|} \right) }
    \nonumber
    \\
    &= \norm{ \left( 1 - \eta \, \alpha \, \xicmin \right)^T
    \left( \theta_0 - \theta^* \right)
    - \eta \, \dpnoise \,
    \sum_{k = 0}^{T - 1} \frac{\left( 1 - \eta \, \alpha \, \xicmin \right)^{k}}{|\cB_{T - k}|} } 
    \label{eq:unrolled-to-thetazero}
    \\
    &\stackrel{d}{=} \norm{ \left( 1 - \eta \, \alpha \, \xicmin \right)^T
    \left( \theta_0 - \theta^* \right)
    + \eta \, \dpnoise \,
    \sum_{k = 0}^{T - 1} \frac{\left( 1 - \eta \, \alpha \, \xicmin \right)^{k}}{|\cB_{T - k}|} } 
    \label{eq:equal-in-distribution}
    \\
    &\leq \left( 1 - \eta \, \alpha \, \xicmin \right)^T
    \norm{ \theta_0 - \theta^* } 
    + \eta \, \sum_{k = 0}^{T - 1} \frac{\left( 1 - \eta \, \alpha \, \xicmin \right)^{k}}{|\cB_{T - k}|} \,
    \blue{\norm{ \dpnoise }}
    \label{eq:traigular-inequality}
    \intertext{Applying Lemma~\ref{lemma:norm-bound}, the right-hand side can be further upper-bounded with probability at least $1 - \delta$, for $\theta$ with $d$ degrees of freedom,}
    &\leq \left( 1 - \eta \, \alpha \, \xicmin \right)^T
    \norm{ \theta_0 - \theta^* } 
    + \eta \, \sum_{k = 0}^{T - 1} \frac{\left( 1 - \eta \, \alpha \, \xicmin \right)^{k}}{|\cB_{T - k}|} \,
    \blue{\sigma \, C \left( \sqrt{d} + \sqrt{2 \log{1 / \delta}} \right)}
    \nonumber
    \\
    &\leq \left( 1 - \eta \, \alpha \, \xicmin \right)^T
    \norm{ \theta_0 - \theta^* } 
    + \eta \, \paren{\sum_{k = 0}^{T - 1} \left( 1 - \eta \, \alpha \, \xicmin \right)^{k}} \,
    \frac{\sigma \, C}{|\cB|_{\min}} \left( \sqrt{d} + \sqrt{2 \log{1 / \delta}} \right)
    \label{batch-min}
    \\
    &= \left( 1 - \eta \, \alpha \, \xicmin \right)^T
    \norm{ \theta_0 - \theta^* } 
    + \frac{1 - \left( 1 - \eta \, \alpha \, \xicmin \right)^{T}}
        {\alpha \, \xicmin} \,
    \frac{\sigma \, C}{|\cB|_{\min}} \left( \sqrt{d} + \sqrt{2 \log{1 / \delta}} \right)
    \label{eq:geometric-series}
    \intertext{Let $\bac = \alpha \xicmin$, $\sfactor = \left( 1 - \left( 1 - \eta \bac \right)^{T} \right) / \bac$ and $\Delta_{d, \delta} = \left( \sqrt{d} + \sqrt{2 \log{1 / \delta}} \right)$; then we have}
    \norm{ \theta_T - \theta^* }
    &\leq \left( 1 - \eta \, \bac \right)^T
    \norm{ \theta_0 - \theta^* } 
    + \frac{\sigma \, C}{|\cB|_{\min}} \, \sfactor \, \Delta_{d, \delta}
    \label{eq:substituting-abvr}
    \\
    \intertext{Multiplying both sides by $-1$ transforms the left-hand side into the collective's success. This converts the upper bound on the parameter norm difference into a lower bound on collective's success, resulting in the final bound:}
    S_T(\alpha, \sigma, C) 
    &\geq -\left( 1 - \eta \, \bac \right)^T
    \norm{ \theta_0 - \theta^* } 
    - \frac{\sigma \, C}{|\cB|_{\min}} \, \sfactor \, \Delta_{d, \delta}.
    \nonumber
\end{align}
\end{proof}

Additional details for some steps of the proof are provided for clarity.
In step \eqref{eq:substitute-redirect}, we substitute the firm's clipped gradient with the collective's gradient-redirecting strategy defined in Equation \eqref{eq:clipped-gradient-redirect}.
In step \eqref{eq:factor-terms}, we factor out $(\theta_{T-1} - \theta^*)$ to isolate the contraction factor.
In step \eqref{eq:relaxed-bound-xicmin}, we substitute $\theta_{T - 1}$ with a smaller value, yielding a relaxed upper bound.
Step \eqref{eq:unrolled-to-thetazero} involves unrolling the gradient-update recursion, which leads to a geometric series whose first term is 1 and common ratio $1 - \eta \, \alpha \, \xicmin$.
In step \eqref{eq:equal-in-distribution}, we use the fact that adding or subtracting a zero-centered Gaussian random variable results in random variables that are equal in distribution.
In step \eqref{eq:traigular-inequality}, we apply the triangle inequality, where the sum of norms is greater than or equal to the norm of the sum.
In step \eqref{batch-min}, we bound the sum by assuming a minimum batch size $|\cB|_{\min}$ for all $T$ steps.
Finally, in step \eqref{eq:geometric-series}, we evaluate the finite geometric series,
noting that the step size $\eta$ cancels out to leave $\alpha \xicmin$ in the denominator.

\subsection{Theorem~\ref{theorem:main} in the No-Privacy Case}
\label{ap-subsec:proofs-reduction}

In this section, we show that, in the \emph{no-privacy} regime, our main bound reduces to the bound stated by \citet[Theorem~10]{hardt2024aca}, which is restated in Theorem~\ref{theorem:hardt-theorem-10}.
We begin with the bound from our main Theorem~\ref{theorem:main}:
\begin{align}
    S_T(\alpha, \sigma, C) 
    &\geq -\left( 1 - \eta \, \bac \right)^T
    \norm{ \theta_0 - \theta^* } 
    - \frac{\sigma \, C}{|\cB|_{\min}} \, \sfactor \, \Delta_{d, \delta}.
    \label{eq:main-bound}
\end{align}
The term $\bac$ is defined as follows from Equation~\eqref{eq:substituting-abvr}:
\begin{align}
    \bac &= \alpha \, \xicmin
    \nonumber
\end{align}
where $\xicmin = \min_{\lambda \in [0, 1]} \xic (\lambda \theta_0 + (1 - \lambda) \theta^*)$.
To better understand the behavior of $\xic$, we focus on its value at a particular point $\theta_t$ along the trajectory, as defined in Equation~\eqref{eq:xic-definition}:
\begin{align}
    \xi^c(\theta_t) 
    &= \frac{\norm{ \gbarclip{\cB_t'} (\theta_t)
    + \tfrac{1 - \alpha}{\alpha}\, \gbarclip{\cB_0} (\theta_t) }}
            {\norm{ \theta_t - \theta^* }}.
    \label{eq:xi-clip-value}
\end{align}

\paragraph{No-privacy Specialization ($\sigma=0$ and $C\to\infty$).}

Setting $\sigma = 0$ eliminates the second term from Equation~\eqref{eq:main-bound}, giving
\begin{align}
    S_{T}(\alpha, C)
    &\geq - \left( 1 - \eta \bac \right)^T \norm{ \theta_0 - \theta^* }
    \label{eq:main-bound-intermediate}
\end{align}
By the definition of clipping operator, as the threshold $C \to \infty$, the clipping becomes inactive. 
Using this fact and considering expectation of unclipped gradient, Equation~\eqref{eq:xi-clip-value} can be written as:
\begin{align}
    \xi(\theta_t) 
    &= \frac{\norm{ \g{\cP'} (\theta_t)
    + \tfrac{1 - \alpha}{\alpha}\, \g{\cP_0} (\theta_t) }}
            {\norm{ \theta_t - \theta^* }}.
\end{align}
This definition of $\xi$ is identical to original Equation~\eqref{eq:original-xi}.
Hence, when $\sigma=0$ and $C\to\infty$, 
\begin{align}
    \bac &= \alpha \, \xicmin
    \nonumber \\
    &= \alpha \, \ximin
    \nonumber \\
    &= C(\alpha) \quad (\text{from Theorem~\ref{theorem:hardt-theorem-10}})
\end{align}
Finally, the Equation~\eqref{eq:main-bound-intermediate} reduces to:
\begin{align}
    S_{T}(\alpha)
    &\geq - \left( 1 - \eta C(\alpha) \right)^T \norm{ \theta_0 - \theta^* }
\end{align}
which matches the original bound of \citet{hardt2024aca}.

\section{Additional Results}
\label{ap-sec:results}

\subsection{Baseline accuracies}
\label{ap-subsec:results-baseline}

Table~\ref{tab:model_acc} presents the baseline predictive accuracies of DP-trained classifiers across all datasets.

\begin{table}[ht]
    \centering
    \begin{tabular}{l|ccc|ccc|ccc|ccc}
        \toprule
        \text{Dataset} 
        & \multicolumn{3}{c|}{\text{\cifarten}} 
        & \multicolumn{3}{c|}{\text{\mnist}} 
        & \multicolumn{3}{c|}{\text{\resume}} 
        & \multicolumn{3}{c}{\text{\bm}} \\
        $C$ & 0.44 & 3.97 & 21.83 & 0.08 & 0.63 & 3.39 & 0.84 & 1.80 & 3.44 & 0.92 & 2.51 & 4.94 \\
        \midrule
        $\varepsilon = 15.0$   & 73\% & 61\% & 48\%  & 91\% & 97\% & 97\%  & 86\% & 86\% & 86\%  & 83\% & 82\% & 82\% \\
        $\varepsilon = 5.0$    & 71\% & 57\% & 44\%  & 90\% & 97\% & 97\%  & 86\% & 86\% & 85\%  & 83\% & 82\% & 82\% \\
        $\varepsilon = 2.5$    & 68\% & 49\% & 40\%  & 88\% & 96\% & 96\%  & 86\% & 86\% & 85\%  & 83\% & 83\% & 81\% \\
        $\varepsilon = 1.5$    & 63\% & 47\% & 34\%  & 85\% & 95\% & 96\%  & 86\% & 86\% & 84\%  & 83\% & 82\% & 80\% \\
        $\varepsilon = 1.25$   & 60\% & 46\% & 33\%  & 83\% & 95\% & 95\%  & 86\% & 86\% & 84\%  & 83\% & 81\% & 79\% \\
        $\varepsilon = 1.0$    & 57\% & 44\% & 29\%  & 81\% & 95\% & 95\%  & 86\% & 86\% & 84\%  & 83\% & 80\% & 74\% \\
        \bottomrule
    \end{tabular}
    \caption[
        Test set accuracies under different privacy configurations
    ]{
        Test set accuracies under different privacy losses ($\varepsilon$) 
        for models trained on four datasets: \cifarten, \mnist, \resume, and \bm.
        The clipping thresholds were selected based on the 25th, 50th, and 75th percentiles of per-sample gradient norms computed over the first few epochs of non-private training~\citep{abadi2016dpsgd}.
        For \cifarten, \mnist, and \bm, the reported metric is classification accuracy, while for \resume, the reported accuracy corresponds to $1 \, -$ Hamming loss for the multi-label classification setting.
        Non-private baselines ($\varepsilon = \infty,\, \sigma = 0,\, C = \infty$) 
        achieve accuracies of 86\%, 100\%, 96\%, and 82\%, respectively.
    }
    \label{tab:model_acc}
\end{table}

\subsection{Critical Mass Plots}
\label{ap-subsec:results-critical}

\begin{figure*}[ht]
    \centering
    \begin{subfigure}[b]{0.98\linewidth}
        \centering
        \includegraphics[width=\linewidth]{plots/critical_mass/max_grad_norm_legend.pdf}
    \end{subfigure}
    \begin{subfigure}[b]{0.48\linewidth}
        \centering
        \begin{subfigure}[b]{0.45\linewidth}
            \includegraphics[width=\linewidth]{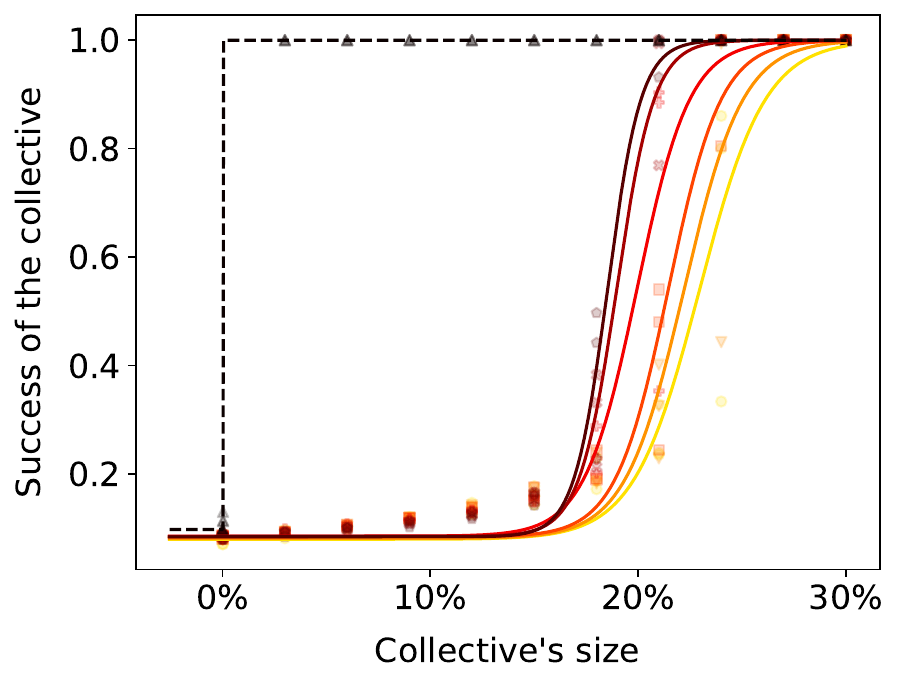}
            \caption*{$C = 0.08$}
        \end{subfigure}
        \hfill
        \begin{subfigure}[b]{0.45\linewidth}
            \includegraphics[width=\linewidth]{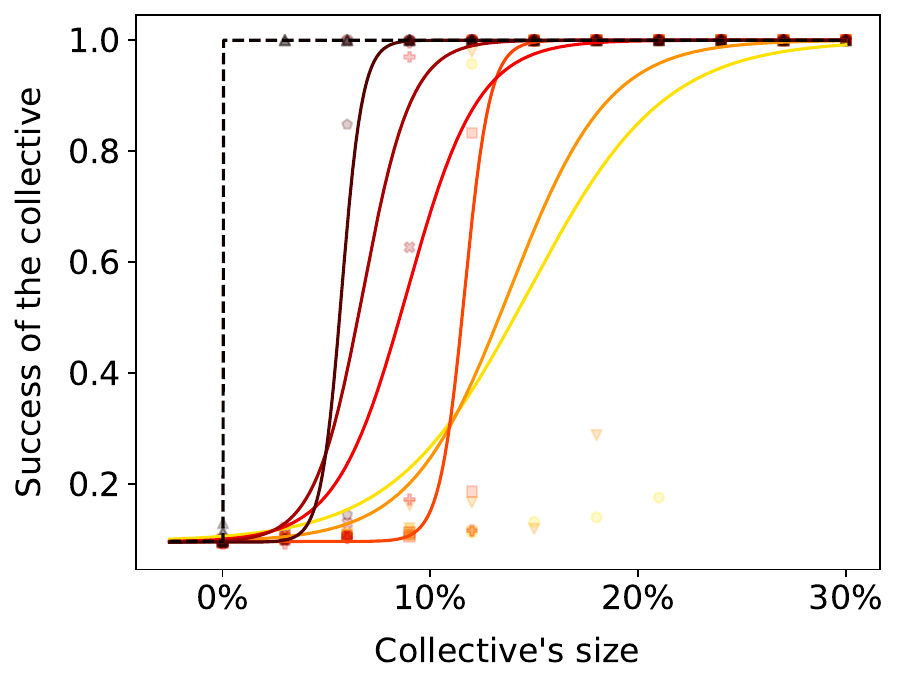}
            \caption*{$C = 3.39$}
        \end{subfigure}
        \caption{\textit{patch} signal with offset 50}
    \end{subfigure}
    \hfill  
    \begin{subfigure}[b]{0.48\linewidth}
        \centering
        \begin{subfigure}[b]{0.45\linewidth}
            \includegraphics[width=\linewidth]{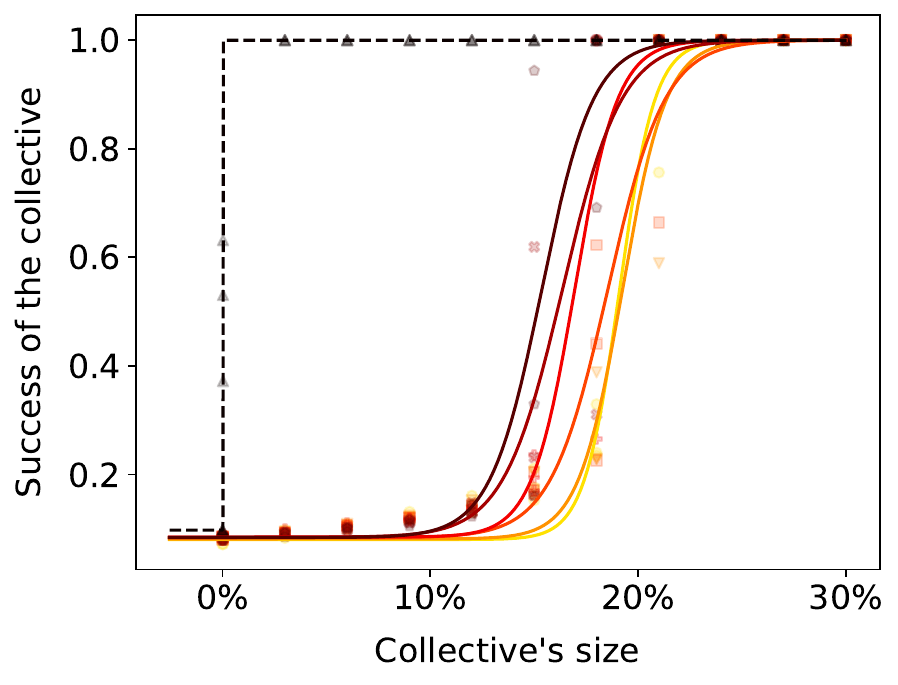}
            \caption*{$C = 0.08$}
        \end{subfigure}
        \hfill
        \begin{subfigure}[b]{0.45\linewidth}
            \includegraphics[width=\linewidth]{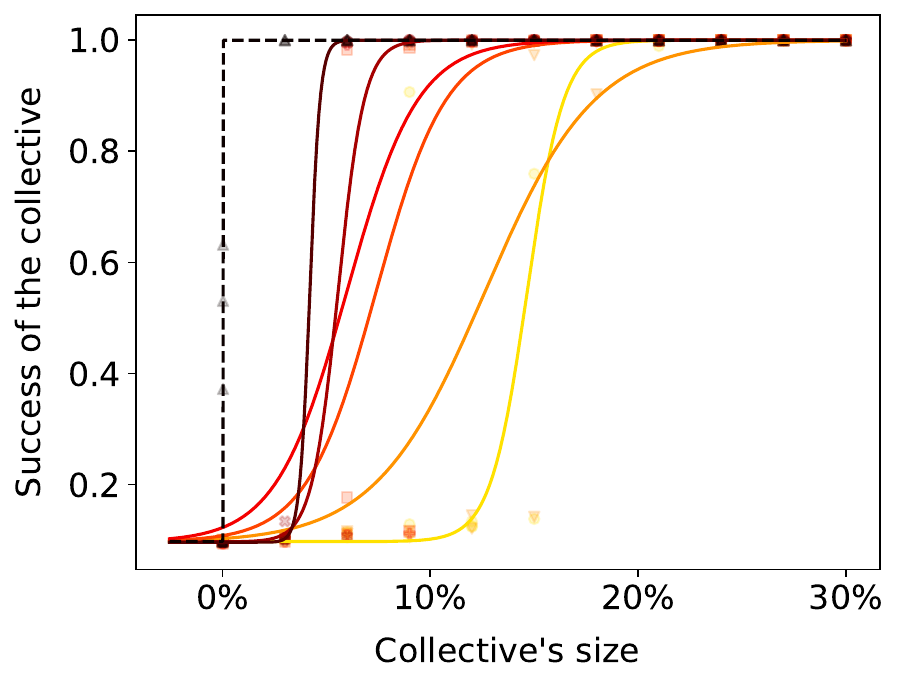}
            \caption*{$C = 3.39$}
        \end{subfigure}
        \caption{\textit{grid} signal with offset 10}
    \end{subfigure}
    \caption[Success of the collective on the \mnist dataset]{
        Success of the collective on the \mnist dataset.
        Each column corresponds to a different clipping threshold $C \in \{0.08, \, 3.39\}$ (as determined by the 25th and 75th percentiles of per-sample gradient norms), and each row to a different collective's strategy.
    }
    \Description{
        Success of the collective across datasets as a function of privacy loss and collective size.
        Results are shown for \mnis comparing different privacy levels to a non-private SGD baseline.
        The horizontal axis shows the collective's size as a percentage of the training data, and the vertical axis shows their success.
        Curves illustrate how stronger privacy constraints require larger collectives to achieve similar performance.
    }
    \label{fig:critical_mass_mnist_rest}
\end{figure*}

\begin{figure*}[ht]
    \centering
    \begin{subfigure}[b]{0.98\linewidth}
        \centering
        \includegraphics[width=\linewidth]{plots/critical_mass/max_grad_norm_legend.pdf}
    \end{subfigure}
    \begin{subfigure}[b]{0.48\linewidth}
        \begin{subfigure}[b]{0.45\linewidth}
            \includegraphics[width=\linewidth]{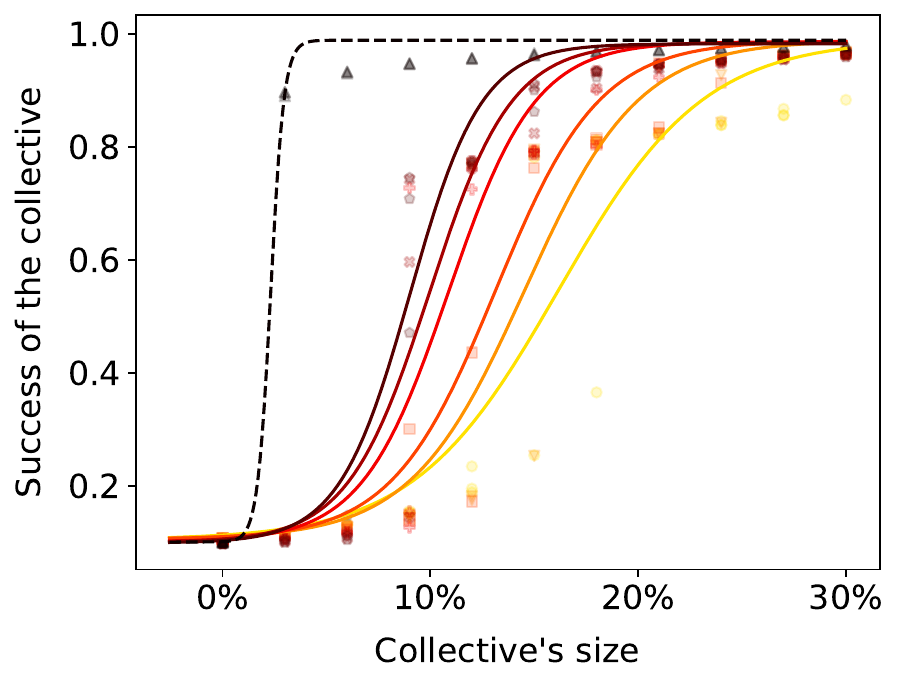}
            \caption*{$C = 0.44$}
        \end{subfigure}
        \hfill
        \begin{subfigure}[b]{0.45\linewidth}
            \includegraphics[width=\linewidth]{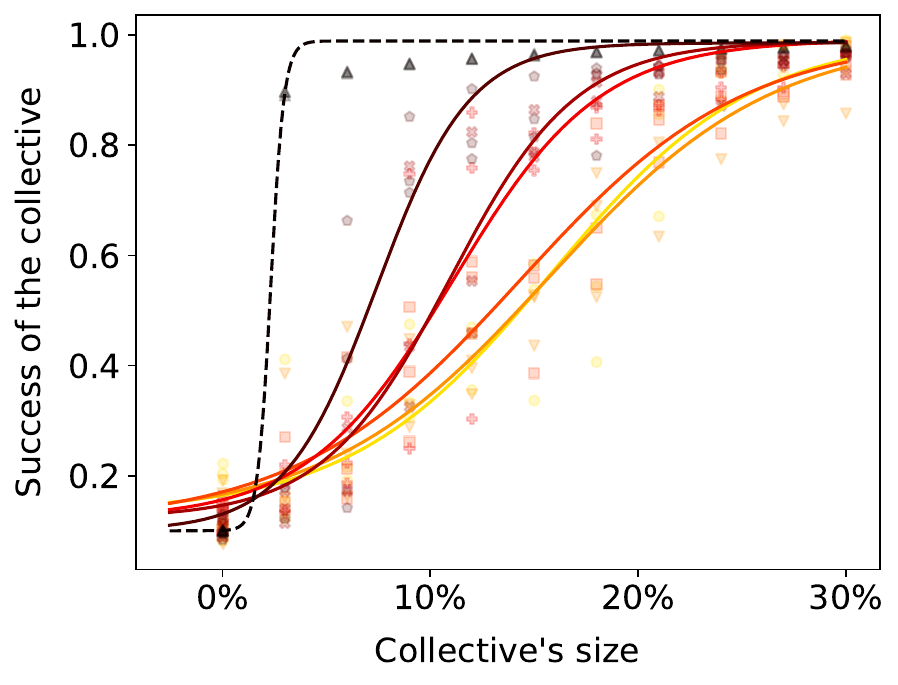}
            \caption*{$C = 21.83$}
        \end{subfigure}
        \caption{\textit{patch} signal with offset 50}
        \label{subfig:critical_mass_cifar10_adaptive_patch}
    \end{subfigure}
    \hfill
    \begin{subfigure}[b]{0.48\linewidth}
        \begin{subfigure}[b]{0.45\linewidth}
            \includegraphics[width=\linewidth]{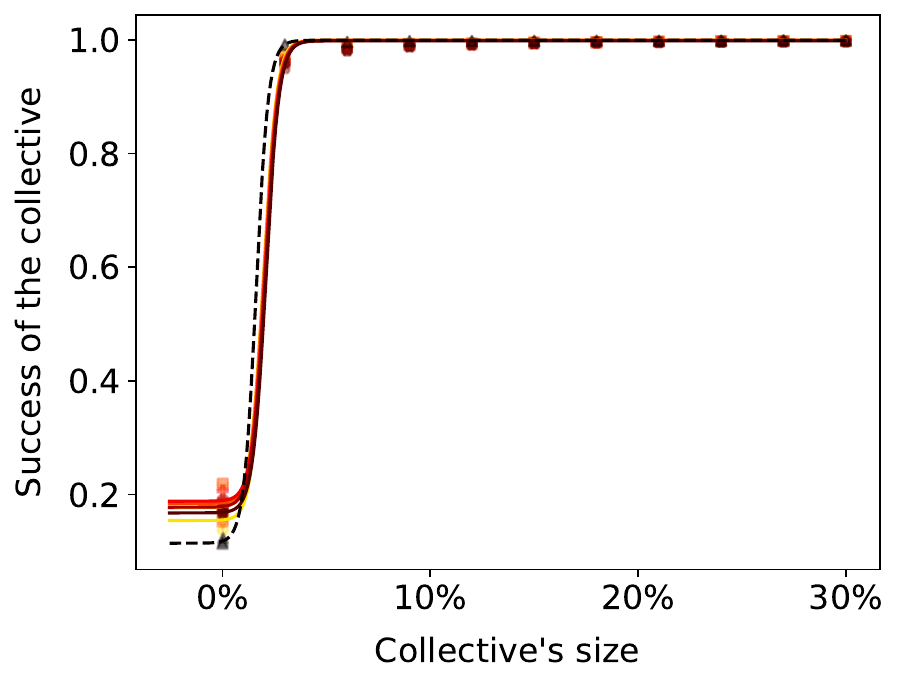}
            \caption*{$C = 0.44$}
        \end{subfigure}
        \hfill
        \begin{subfigure}[b]{0.45\linewidth}
            \includegraphics[width=\linewidth]{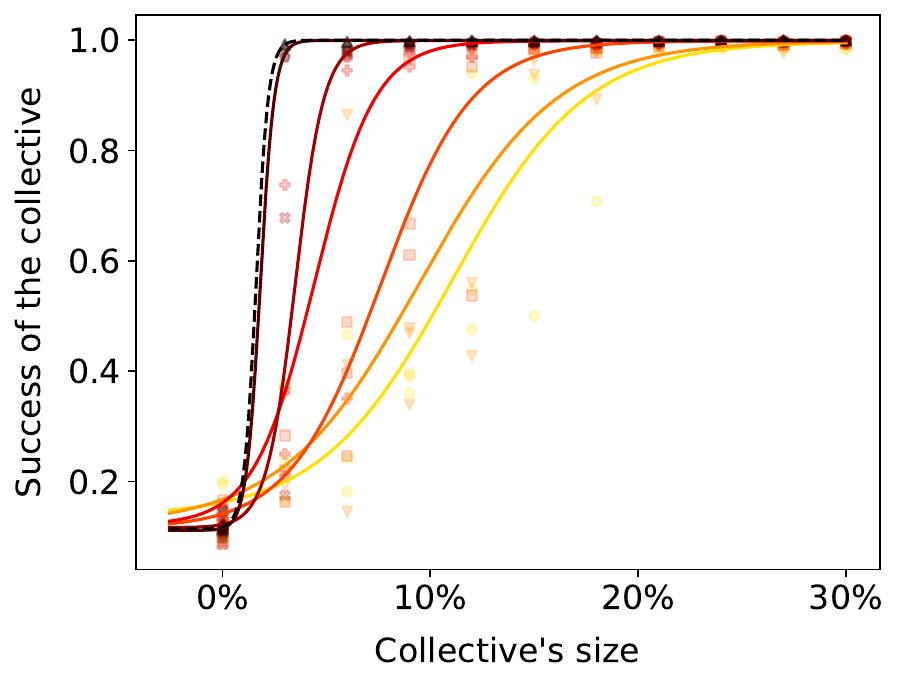}
            \caption*{$C = 21.83$}
        \end{subfigure}
        \caption{\textit{grid} signal with offset 10}
        \label{subfig:critical_mass_cifar10_adaptive_grid}
    \end{subfigure}
    \caption[
        Success of the collective on the \cifarten dataset
    ]{
        Success of the collective on the \cifarten dataset.
        Each column corresponds to a different clipping threshold $C \in \{0.44, \, 21.83\}$ (as determined by the 25th and 50th percentiles of per-sample gradient norms), and each row to a different collective's strategy.
    }
    \Description{
        Success of the collective across datasets as a function of privacy loss and collective size.
        Results are shown for \cifarten comparing different privacy levels to a non-private SGD baseline.
        The horizontal axis shows the collective's size as a percentage of the training data, and the vertical axis shows their success.
        Curves illustrate how stronger privacy constraints require larger collectives to achieve similar performance.
    }
    \label{fig:critical_mass_cifar10_rest}
\end{figure*}

\begin{figure}[ht]
    \centering
    \begin{subfigure}[b]{0.95\linewidth}
        \centering
        \includegraphics[width=\linewidth]{plots/critical_mass/max_grad_norm_legend.pdf}
    \end{subfigure}
    \begin{subfigure}[b]{0.48\linewidth}
        \begin{subfigure}[b]{0.45\linewidth}
            \includegraphics[width=\linewidth]{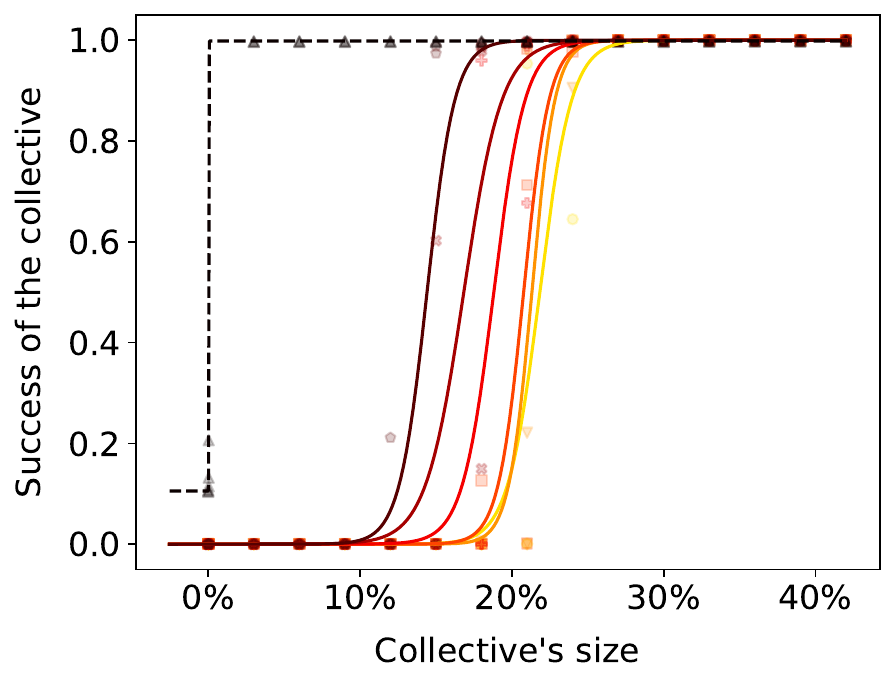}
            \caption*{$C = 0.84$}
        \end{subfigure}
        \hfill
        \begin{subfigure}[b]{0.45\linewidth}
            \includegraphics[width=\linewidth]{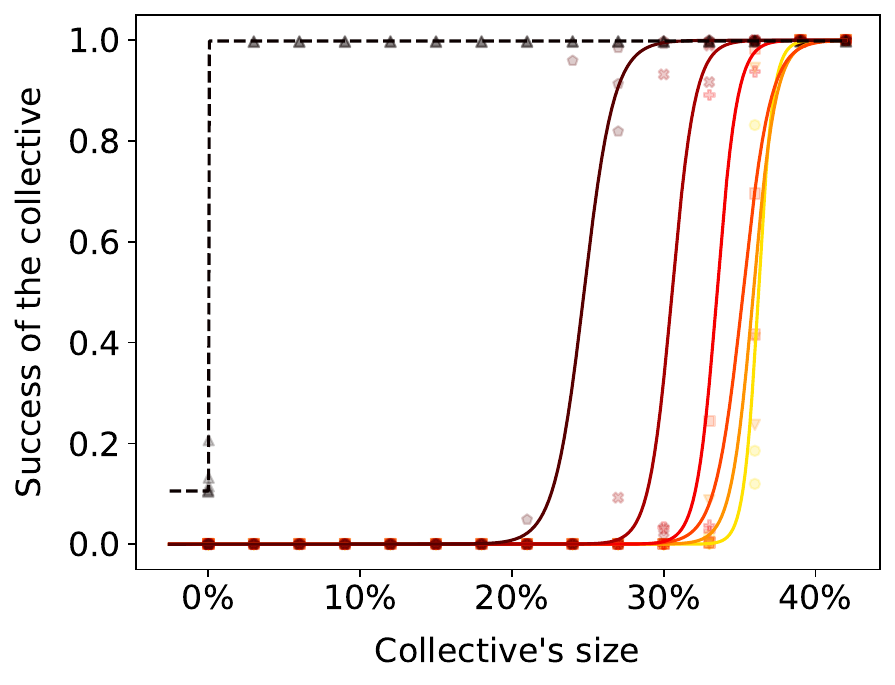}
            \caption*{$C = 3.44$}
        \end{subfigure}
        \caption{\resume dataset}
    \end{subfigure}
    \hfill
    \begin{subfigure}[b]{0.48\linewidth}
        \begin{subfigure}[b]{0.45\linewidth}
            \includegraphics[width=\linewidth]{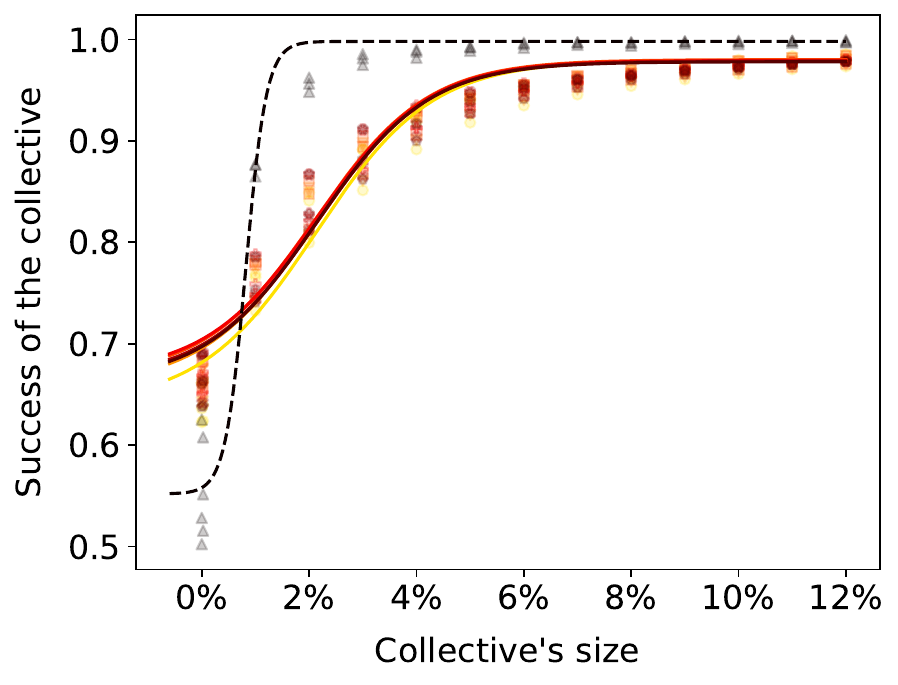}
            \caption*{$C = 0.92$}
            \label{subfig:bm_crit_mass_a}
        \end{subfigure}
        \hfill
        \begin{subfigure}[b]{0.45\linewidth}
            \includegraphics[width=\linewidth]{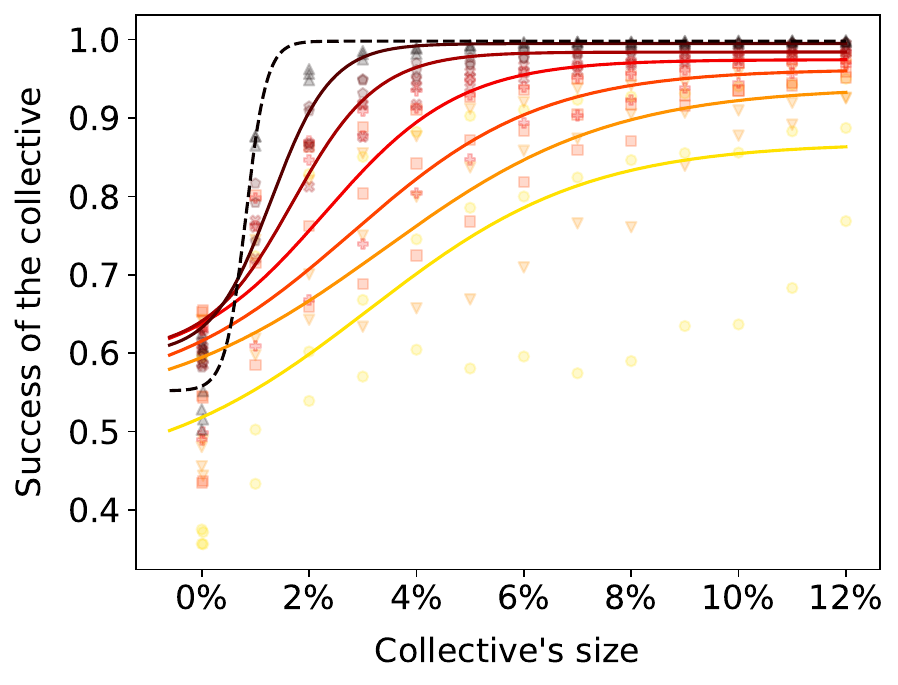}
            \caption*{$C = 4.94$}
            \label{subfig:bm_crit_mass_c}
        \end{subfigure}
        \caption{\bm dataset}
    \end{subfigure}
    \caption[
        Success of the collective on the \resume and \bm datasets.
    ]{
        Success of the collective on the \resume and \bm datasets.
        For \resume dataset, we use $C \in \curly{0.84, \, 3.44}$, and for \bm dataset we use $C \in \curly{0.92, \, 4.94}$ (as determined by the 25th and 75th percentiles of per-sample gradient norms).
    }
    \Description{
        Success of the collective across datasets as a function of privacy loss and collective size.
        Results are shown for \resume, and \bm, comparing different privacy levels to a non-private SGD baseline.
        The horizontal axis shows the collective's size as a percentage of the training data, and the vertical axis shows their success.
        Curves illustrate how stronger privacy constraints require larger collectives to achieve similar performance.
    }
    \label{fig:critical_mass_bm_resume_rest}
\end{figure}

The main paper reports results where the clipping threshold $C$ is selected at the 50th percentile of the per-sample gradient norms, estimated using a non-private learning algorithm trained for a few epochs (see Figure~\ref{fig:gradients}).
Results for the 25th and 75th percentiles are shown in Figures~\ref{fig:critical_mass_mnist_rest} (\mnist),~\ref{fig:critical_mass_cifar10_rest} (\cifarten), and~\ref{fig:critical_mass_bm_resume_rest} (\bm and \resume).


\begin{figure*}[ht]
    \centering
    \begin{subfigure}[b]{0.98\linewidth}
        \centering
        \includegraphics[width=\linewidth]{plots/critical_mass/max_grad_norm_legend.pdf}
    \end{subfigure}
    \begin{subfigure}[b]{0.48\linewidth}
        \centering
        \begin{subfigure}[b]{0.45\linewidth}
            \includegraphics[width=\linewidth]{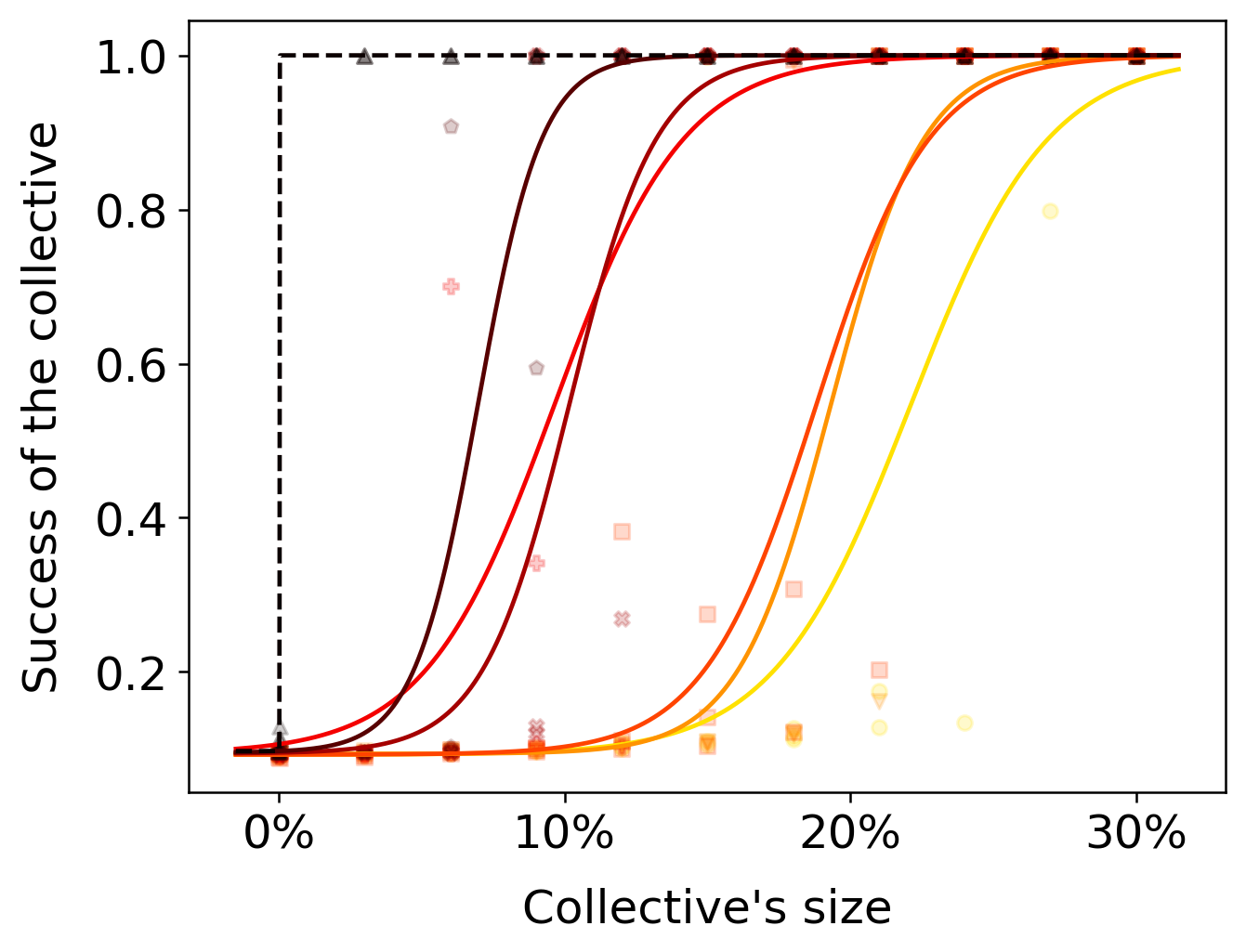}
            \caption*{\textit{patch} with offset 50}
        \end{subfigure}
        \hfill
        \begin{subfigure}[b]{0.45\linewidth}
            \includegraphics[width=\linewidth]{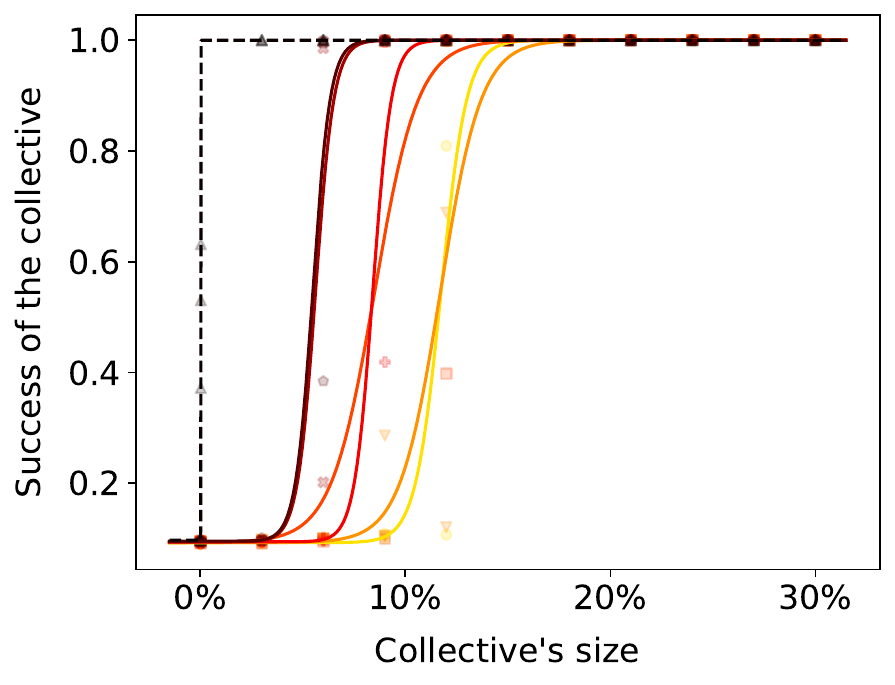}
            \caption*{\textit{grid} with offset 10}
        \end{subfigure}
        \caption{\mnist dataset}
    \end{subfigure}
    \hfill
    \begin{subfigure}[b]{0.48\linewidth}
        \centering
        \begin{subfigure}[b]{0.45\linewidth}
            \includegraphics[width=\linewidth]{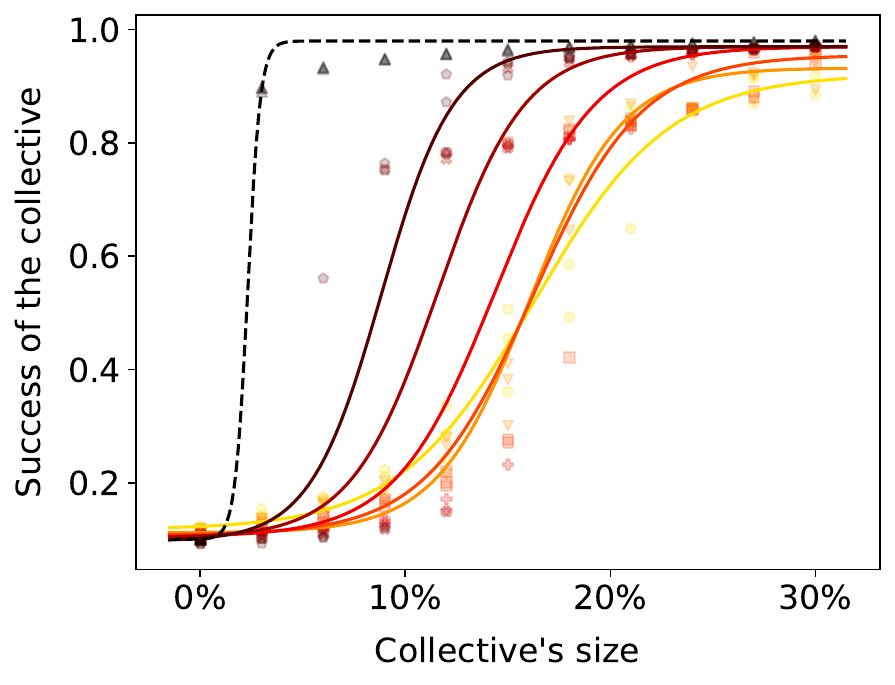}
            \caption*{\textit{patch} with offset 50}
        \end{subfigure}
        \hfill
        \begin{subfigure}[b]{0.45\linewidth}
            \includegraphics[width=\linewidth]{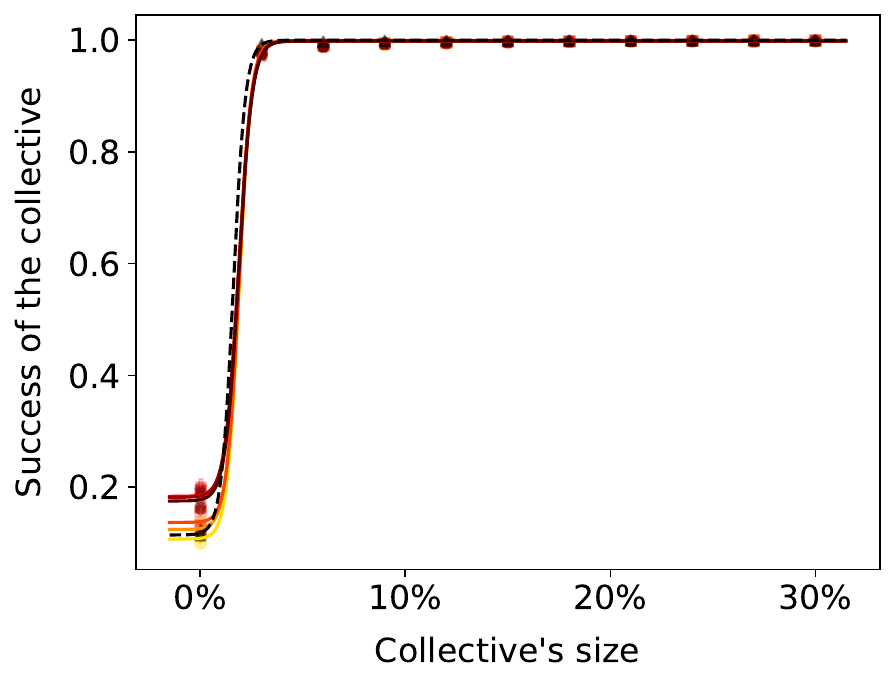}
            \caption*{\textit{grid} with offset 10}
        \end{subfigure}
        \caption{\cifarten dataset}
    \end{subfigure}
    \caption[
        Success of the collective with automatic clipping
    ]{
        Success of the collective with automatic clipping~\cite{bu2023automaticclippingdifferentiallyprivate}.
        Each row corresponds to a different datasets (\mnist and \cifarten), and each column to a different collective's strategy.
    }
    \label{fig:critical-mass-automatic}
\end{figure*}

In addition to the experiments with a fixed clipping threshold, we also evaluate an \emph{automatic clipping} method~\citep{bu2023automaticclippingdifferentiallyprivate}.
This approach automatically adapts the clipping threshold by rescaling each per-sample gradient by its own magnitude ($g_i \leftarrow / \paren{\norm{g_i} + \gamma}$, for $\gamma >0$), then aggregates and adds Gaussian noise.
To implement automatic clipping, we modified the Opacus library according to the code changes outlined in \citet{bu2023automaticclippingdifferentiallyprivate}.
A significant advantage of this method is that it removes the burden on the firm deploying the private model to select and tune the clipping threshold $C$.
With this technique, we observe the same trends, where the critical mass $\alpha^*$ increases as the privacy loss $\varepsilon$ tightens.
The results are shown in Figure~\ref{fig:critical-mass-automatic}.

\subsection{Collective's Success vs. Platform's Accuracy}
\label{ap-subsec:accuracy}

\begin{figure}[ht]
    \centering
    \begin{subfigure}[t]{0.32\linewidth}
        \includegraphics[width=\linewidth]{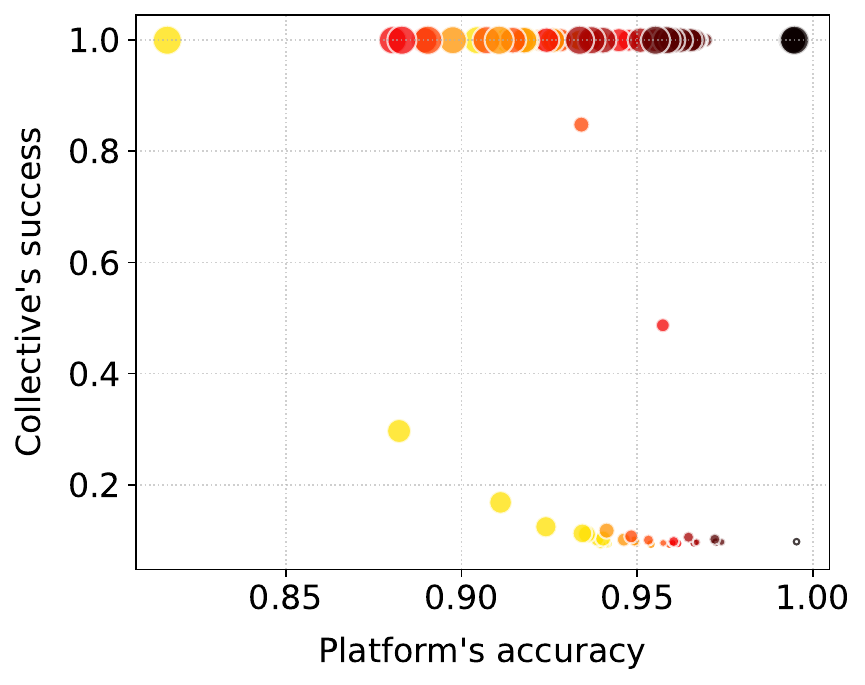}
        \caption{\mnist, patch, $C = 0.63$}
    \end{subfigure}
    \hfill
    \begin{subfigure}[t]{0.32\linewidth}
        \includegraphics[width=\linewidth]{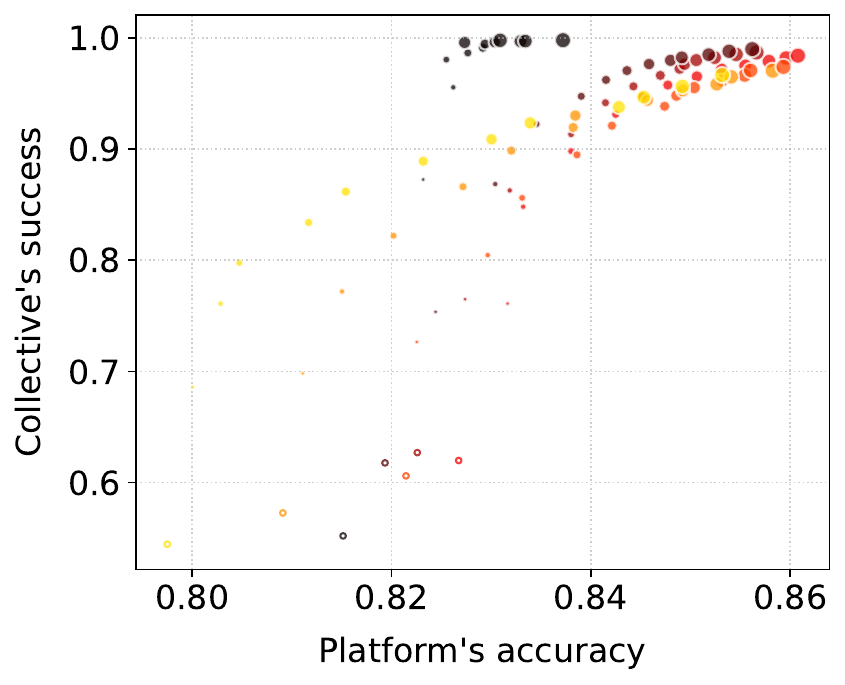}
        \caption{\bm, $C = 2.51$}
    \end{subfigure}
    \hfill
    \raisebox{0.4cm}{
    \begin{subfigure}[t]{0.22\linewidth}
        \includegraphics[width=\linewidth,]{plots/accuracy/cifar10_patch_offset50_legend-max_grad_norm3.97.pdf}
    \end{subfigure}
    }
    \caption[
        Collective's Success vs. Platform's Accuracy under Differential Privacy for \mnist and \bm dataset.
    ]{
        Collective's Success vs. Platform's Accuracy under Differential Privacy for \mnist and \bm dataset.
        Points are colored by the privacy loss ($\varepsilon$) and sized according to the percentage of the dataset controlled by the collective.
        This demonstrates that firms cannot rely on privacy mechanisms to prevent significant influence from a collective without rendering their own models uncompetitive.
    }
    \Description{
        A scatter plot showing the relationship between collective's success and platform's accuracy under differential privacy for \mnist and \bm dataset.
        Point color indicates the level of privacy loss (epsilon), and point size reflects the percentage of the dataset controlled by the collective.
        The plot illustrates that increasing privacy protection does not eliminate collective's influence without also reducing overall model accuracy.
    }
    \label{fig:accuracy_rest}
\end{figure}
Bank Marketing shows an opposite trend compared to the rest of the dataset when it comes to non-private models where the accuracy of the platform after \dpx is improved. We hypothesize that this could be due to added noise during training, improving the generalization of the model, and preventing overfitting.  Similar results can also be seen in \cite{otoumDifferentialPrivacyDrivenFramework2025} where \dpsgd is used for tabular datasets.

\section{Visualization}
\label{ap-sec:visualization}

\begin{figure}[ht]
    \centering
    \begin{subfigure}[b]{\linewidth}
        \centering
        \begin{subfigure}[b]{0.11\linewidth}
            \includegraphics[width=\linewidth]{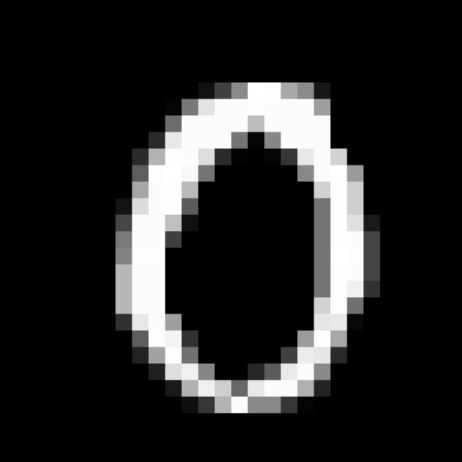}
        \end{subfigure}\hfill
        \begin{subfigure}[b]{0.11\linewidth}
            \includegraphics[width=\linewidth]{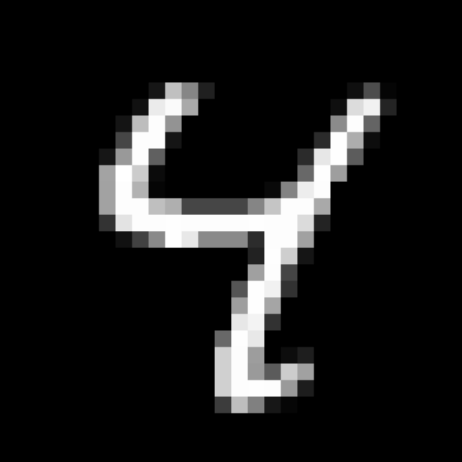}
        \end{subfigure}\hfill
        \begin{subfigure}[b]{0.11\linewidth}
            \includegraphics[width=\linewidth]{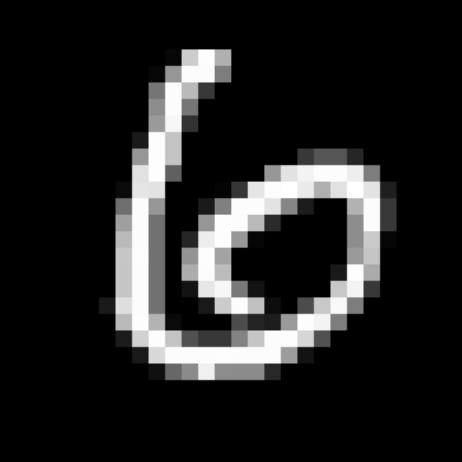}
        \end{subfigure}\hfill
        \begin{subfigure}[b]{0.11\linewidth}
            \includegraphics[width=\linewidth]{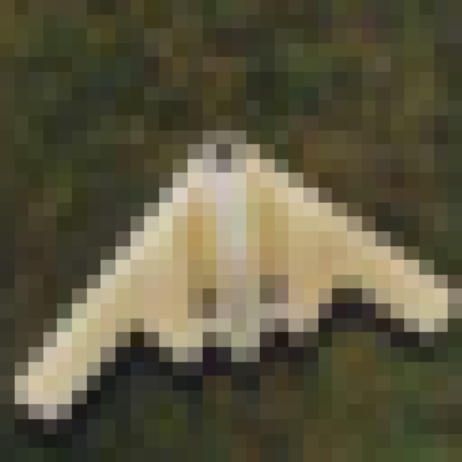}
        \end{subfigure}\hfill
        \begin{subfigure}[b]{0.11\linewidth}
            \includegraphics[width=\linewidth]{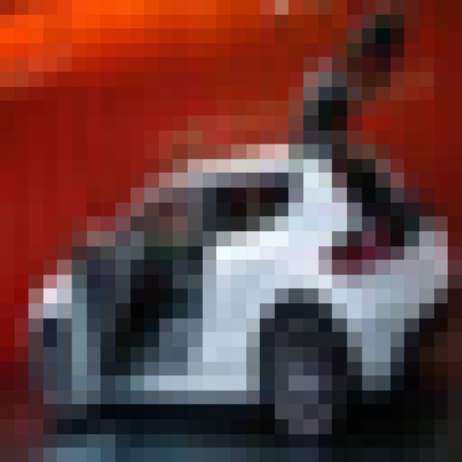}
        \end{subfigure}\hfill
        \begin{subfigure}[b]{0.11\linewidth}
            \includegraphics[width=\linewidth]{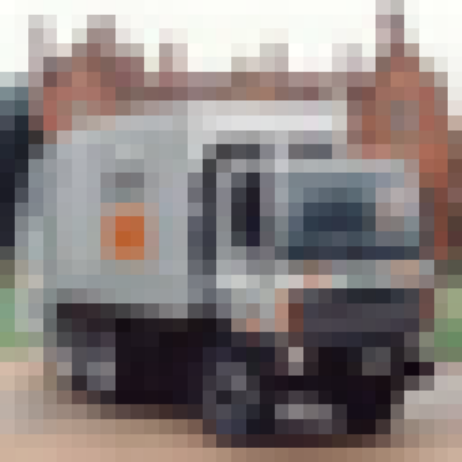}
        \end{subfigure}
        \caption{Original samples from \mnist and \cifarten.}
    \end{subfigure}
    \begin{subfigure}[b]{\linewidth}
        \centering
        \begin{subfigure}[b]{0.11\linewidth}
            \includegraphics[width=\linewidth]{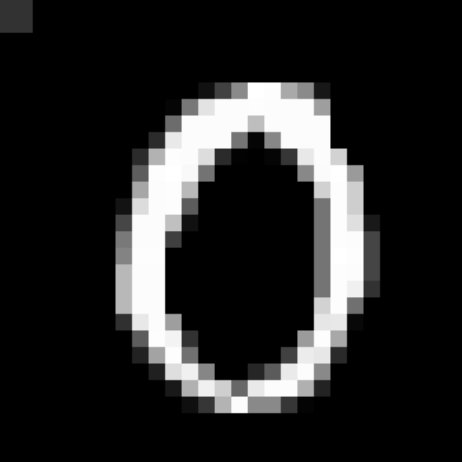}
        \end{subfigure}\hfill
        \begin{subfigure}[b]{0.11\linewidth}
            \includegraphics[width=\linewidth]{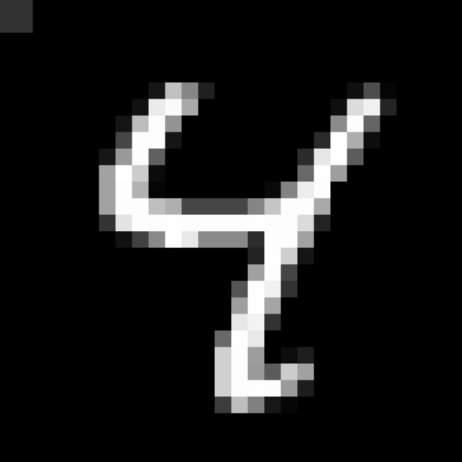}
        \end{subfigure}\hfill
        \begin{subfigure}[b]{0.11\linewidth}
            \includegraphics[width=\linewidth]{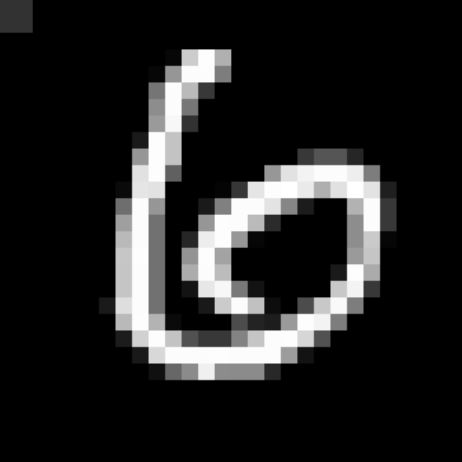}
        \end{subfigure}\hfill
        \begin{subfigure}[b]{0.11\linewidth}
            \includegraphics[width=\linewidth]{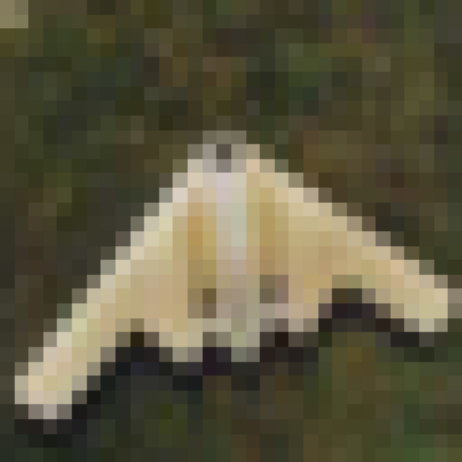}
        \end{subfigure}\hfill
        \begin{subfigure}[b]{0.11\linewidth}
            \includegraphics[width=\linewidth]{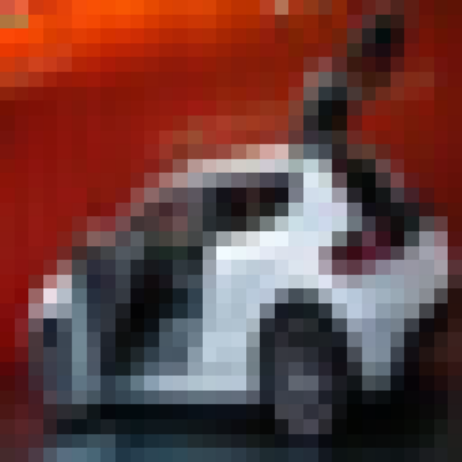}
        \end{subfigure}\hfill
        \begin{subfigure}[b]{0.11\linewidth}
            \includegraphics[width=\linewidth]{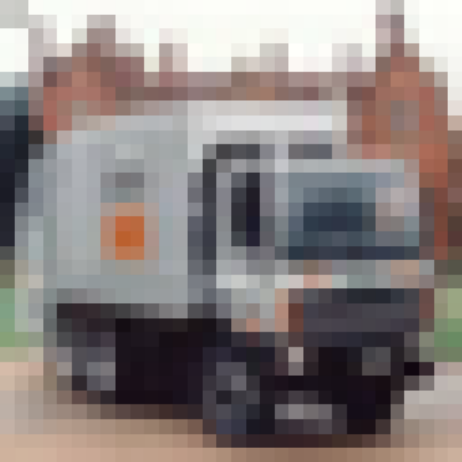}
        \end{subfigure}
        \caption{Samples with \emph{patch} signal applied.}
    \end{subfigure}
    \begin{subfigure}[b]{\linewidth}
        \centering
        \begin{subfigure}[b]{0.11\linewidth}
            \includegraphics[width=\linewidth]{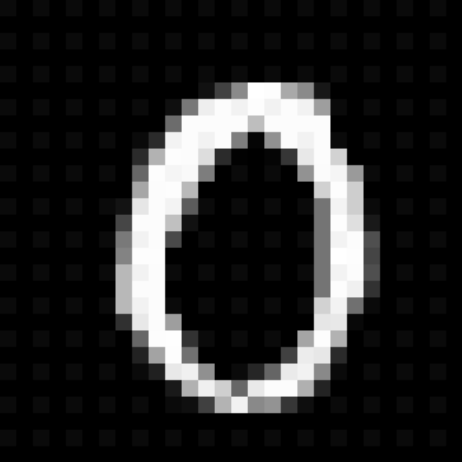}
        \end{subfigure}\hfill
        \begin{subfigure}[b]{0.11\linewidth}
            \includegraphics[width=\linewidth]{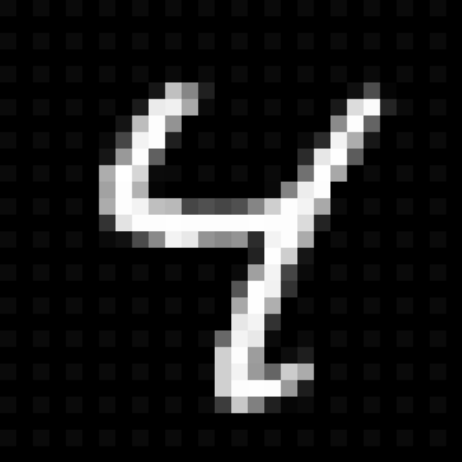}
        \end{subfigure}\hfill
        \begin{subfigure}[b]{0.11\linewidth}
            \includegraphics[width=\linewidth]{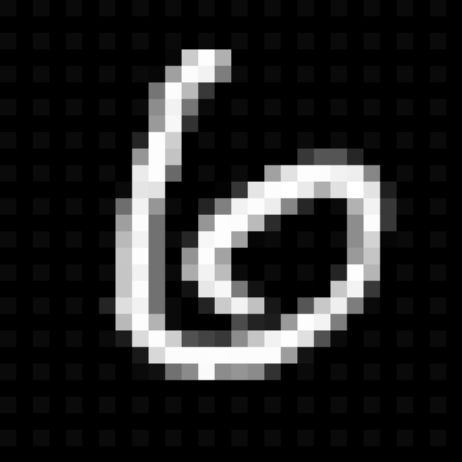}
        \end{subfigure}\hfill
        \begin{subfigure}[b]{0.11\linewidth}
            \includegraphics[width=\linewidth]{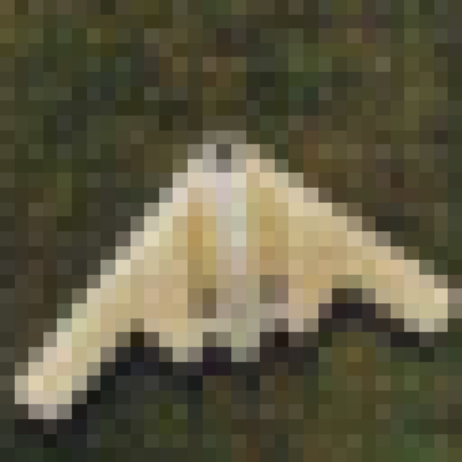}
        \end{subfigure}\hfill
        \begin{subfigure}[b]{0.11\linewidth}
            \includegraphics[width=\linewidth]{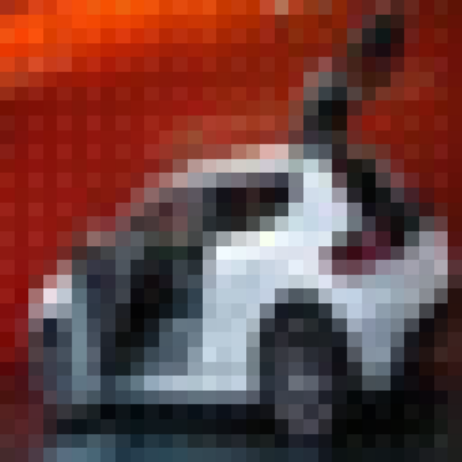}
        \end{subfigure}\hfill
        \begin{subfigure}[b]{0.11\linewidth}
            \includegraphics[width=\linewidth]{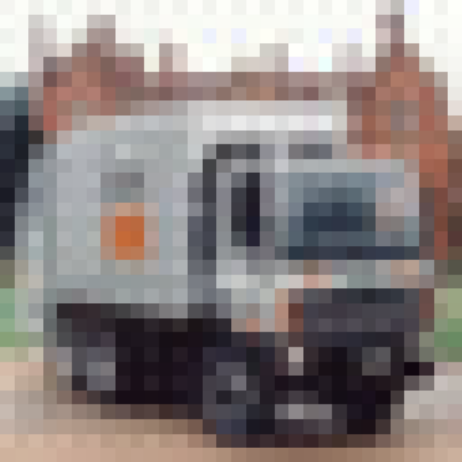}
        \end{subfigure}
        \caption{Samples with \emph{grid} signal applied.}
    \end{subfigure}
    \caption[
        Data points with and without application of signal
    ]{
        Comparison of original, \emph{patch}-signaled, and \emph{grid}-signaled samples for \mnist and \cifarten.
    }
    \Description{
        This figure shows random sample drawn from \mnist and \cifarten with and without the application of \emph{patch} and \emph{grid} signals.
    }
    \label{fig:mnist_cifar_samples}
\end{figure}

Figure~\ref{fig:mnist_cifar_samples} shows the examples of the image before and after the signal is applied for image datasets.

\begin{figure}[ht]
    \centering
    \begin{subfigure}[b]{\linewidth}
        \centering
        \begin{subfigure}[b]{0.34\linewidth}
            \includegraphics[width=\linewidth]{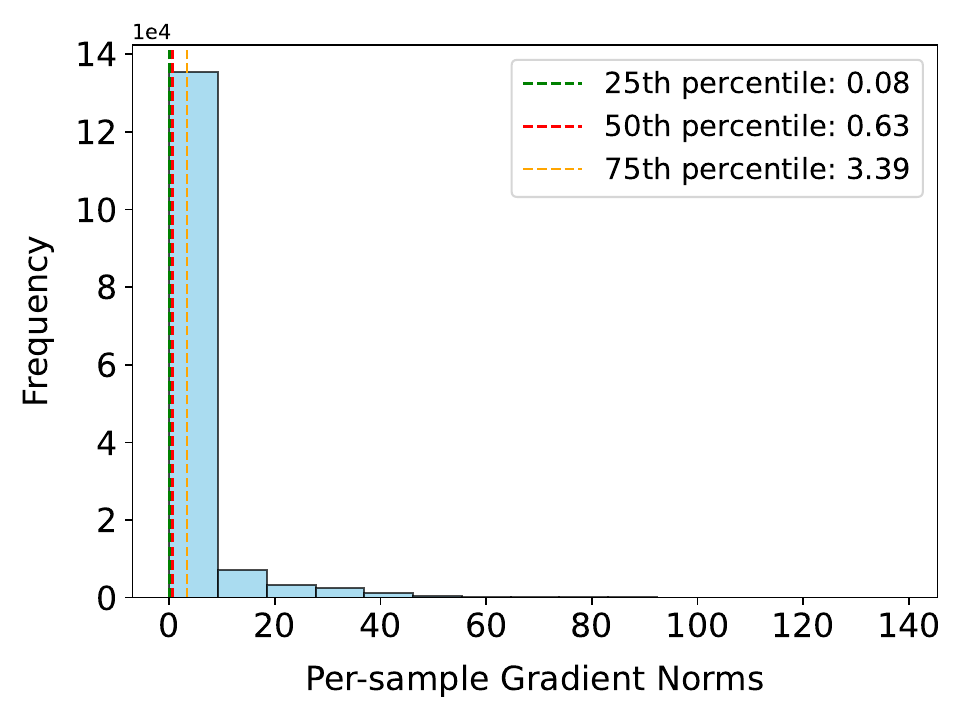}
            \caption{\mnist}
        \end{subfigure}
        \begin{subfigure}[b]{0.34\linewidth}
            \includegraphics[width=\linewidth]{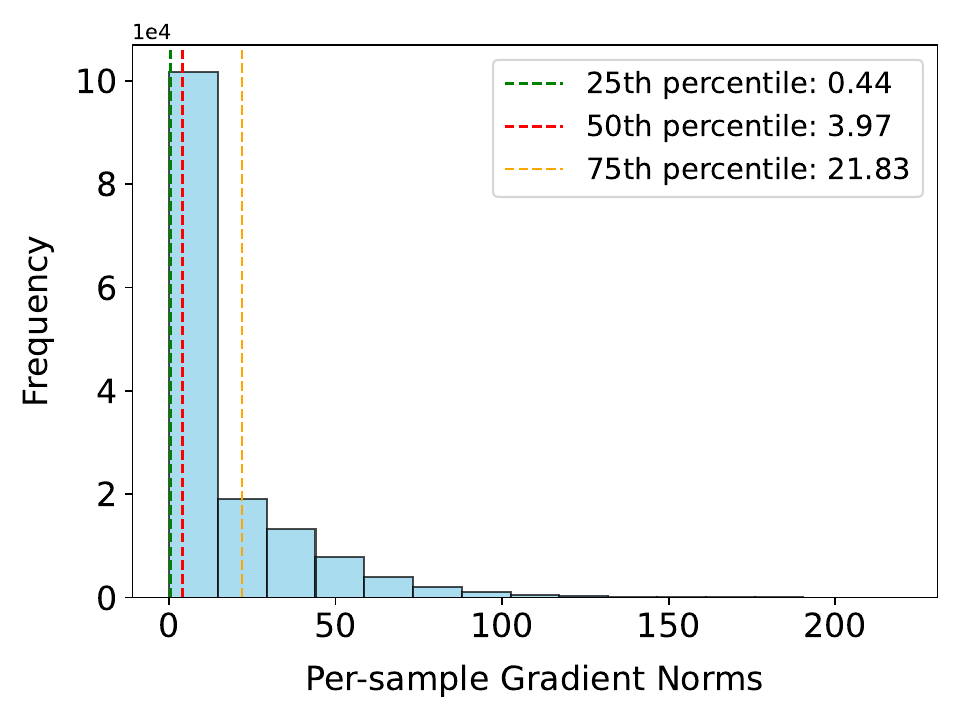}
            \caption{\cifarten}
        \end{subfigure}
        \\
        \begin{subfigure}[b]{0.34\linewidth}
            \includegraphics[width=\linewidth]{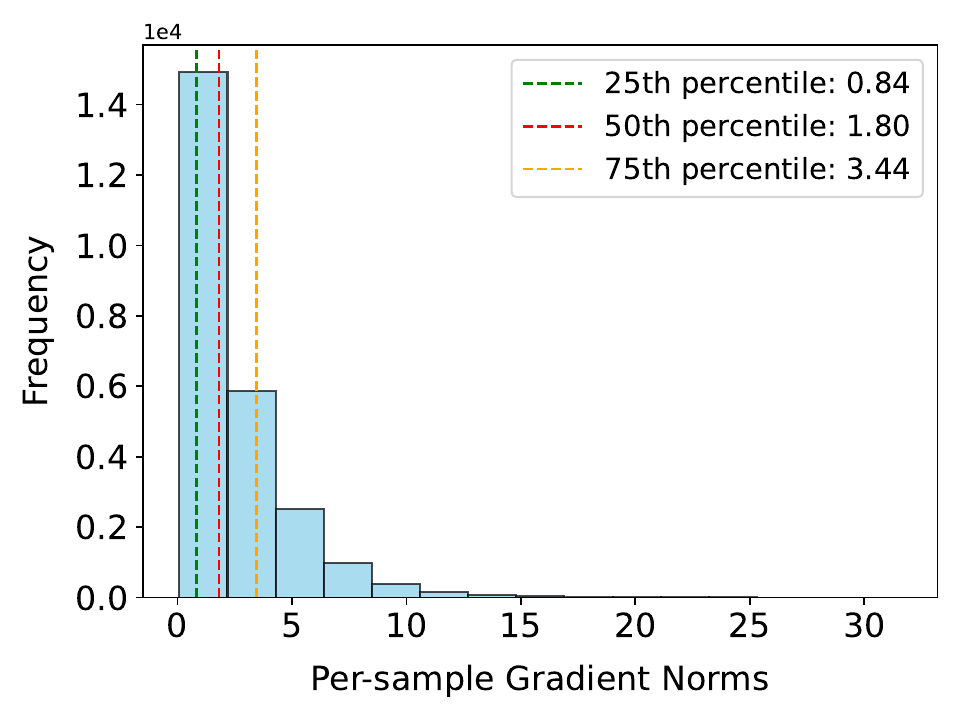}
            \caption{\resume}
        \end{subfigure}
        \begin{subfigure}[b]{0.34\linewidth}
            \includegraphics[width=\linewidth]{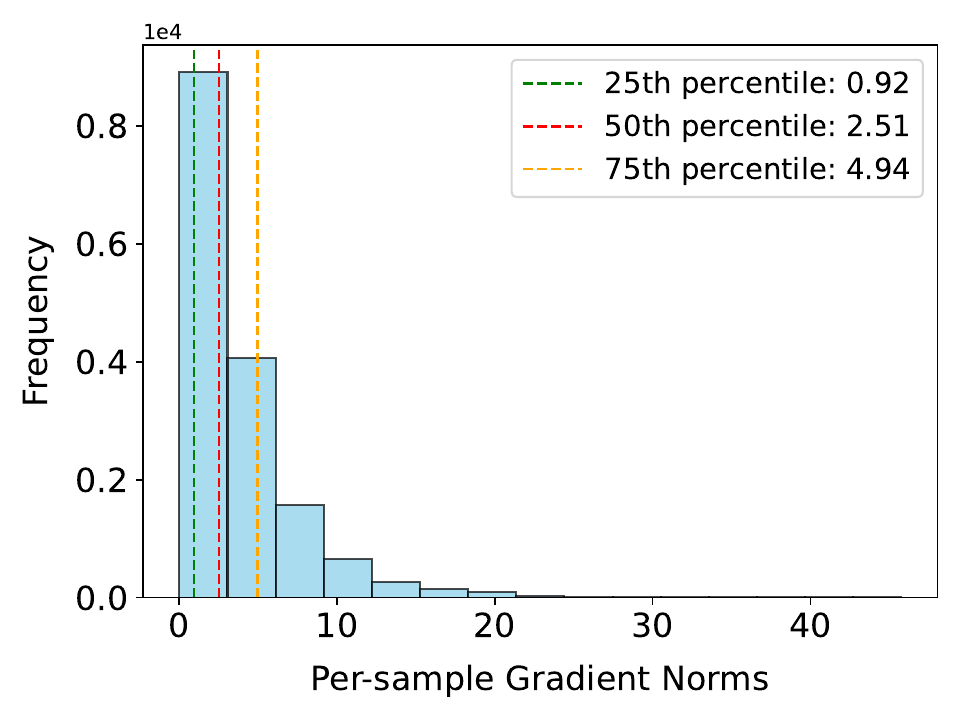}
            \caption{\bm}
        \end{subfigure}
    \end{subfigure}
    \caption[
        Histograms of per-sample gradient norms
    ]{
        Histograms of per-sample gradient norms across all datasets: \mnist, \cifarten, \resume, and \bm dataset.
        Sample gradients were collected during the first few training epochs without applying \dpx constraints.
    }
    \Description{
        This figure shows histogram of per-sample gradient norms across all datasets: \mnist, \cifarten, \resume, and \bm dataset.
        Sample gradients were collected during the first few training epochs without applying \dpx constraints.
    }
    \label{fig:gradients}
\end{figure}

Figure~\ref{fig:gradients} shows the empirical distributions of per-sample gradient norms used to select the clipping thresholds for each dataset.

\end{document}